%% file: submit_aaai.tex
\def\year{2021}\relax
\documentclass[letterpaper]{article}
\usepackage{aaai21} % DO NOT CHANGE THIS
\usepackage{times} % DO NOT CHANGE THIS
\usepackage{helvet} % DO NOT CHANGE THIS
\usepackage{courier} % DO NOT CHANGE THIS
\usepackage[hyphens]{url} % DO NOT CHANGE THIS
\usepackage{graphicx} % DO NOT CHANGE THIS
\usepackage{graphicx}  % DO NOT CHANGE THIS
\usepackage{natbib}
\usepackage{caption}
\usepackage{booktabs}
\usepackage{multirow}
\usepackage{amsmath, amsfonts}
\usepackage[table]{xcolor}
\usepackage{float}
\definecolor{Gray}{gray}{0.93}

\usepackage{soul}

\DeclareMathOperator*{\argmin}{arg\,min}

\usepackage{comment}

\newcommand{\xiyin}[1]{\textcolor{black}{#1}}

\newcommand{\leizhang}[1]{\textcolor{black}{#1}}
\newcommand{\kevin}[1]{\textcolor{black}{#1}}

\newcommand{\Paragraph}[1]{\vspace{1mm} \noindent \textbf{#1} \hspace{0mm}}

\newcommand{\ie}{\textit{i.e.}}
\newcommand{\eg}{\textit{e.g.}}

\def\tabvspace{{\vspace{-4mm}}}
\def\figvspace{{\vspace{-4mm}}}
\makeatletter
  \newcommand\figcaption{\def\@captype{figure}\caption}
  \newcommand\tabcaption{\def\@captype{table}\caption}
\makeatother

\newenvironment{tight_itemize}{
\begin{itemize}
  \setlength{\topsep}{0pt}
  \setlength{\itemsep}{2pt}
  \setlength{\parskip}{0pt}
  \setlength{\parsep}{0pt}
}{\end{itemize}}

\title{VIVO: Visual Vocabulary Pre-Training for Novel Object Captioning}
\author{
Xiaowei Hu,
Xi Yin, 
Kevin Lin, 
Lijuan Wang, 
Lei Zhang, \\
Jianfeng Gao, 
Zicheng Liu \\
}
\affiliations {
    Microsoft Corporation \\
 \{xiaowh, keli, lijuanw, leizhang, jfgao, zliu\}@microsoft.com,
 yinxi.whu@gmail.com
}

\begin{document}

\maketitle

\begin{abstract}

It is highly desirable yet challenging to generate image captions that can describe novel objects which are unseen in caption-labeled training data, a capability that is evaluated in the novel object captioning challenge (nocaps). In this challenge, no additional image-caption training data, other than COCO Captions, is allowed for model training. Thus, conventional Vision-Language Pre-training (VLP) methods cannot be applied. This paper presents VIsual VOcabulary pre-training (VIVO) that performs pre-training in the absence of caption annotations. By breaking the dependency of paired image-caption training data in VLP, VIVO can leverage large amounts of paired image-tag data to learn a visual vocabulary. This is done by pre-training a multi-layer Transformer model that learns to align image-level tags with their corresponding image region features. To address the unordered nature of image tags, VIVO uses a Hungarian matching loss with masked tag prediction to conduct pre-training.

We validate the effectiveness of VIVO by fine-tuning the pre-trained model for image captioning. In addition, we perform an analysis of the visual-text alignment inferred by our model. The results show that our model can not only generate fluent image captions that describe novel objects, but also identify the locations of these objects. Our single model has achieved new state-of-the-art results on nocaps and surpassed the human CIDEr score.
\end{abstract}

\section{Introduction}
\begin{figure}[t]
% \begin{center}
\centering
\includegraphics[width=0.46\textwidth]{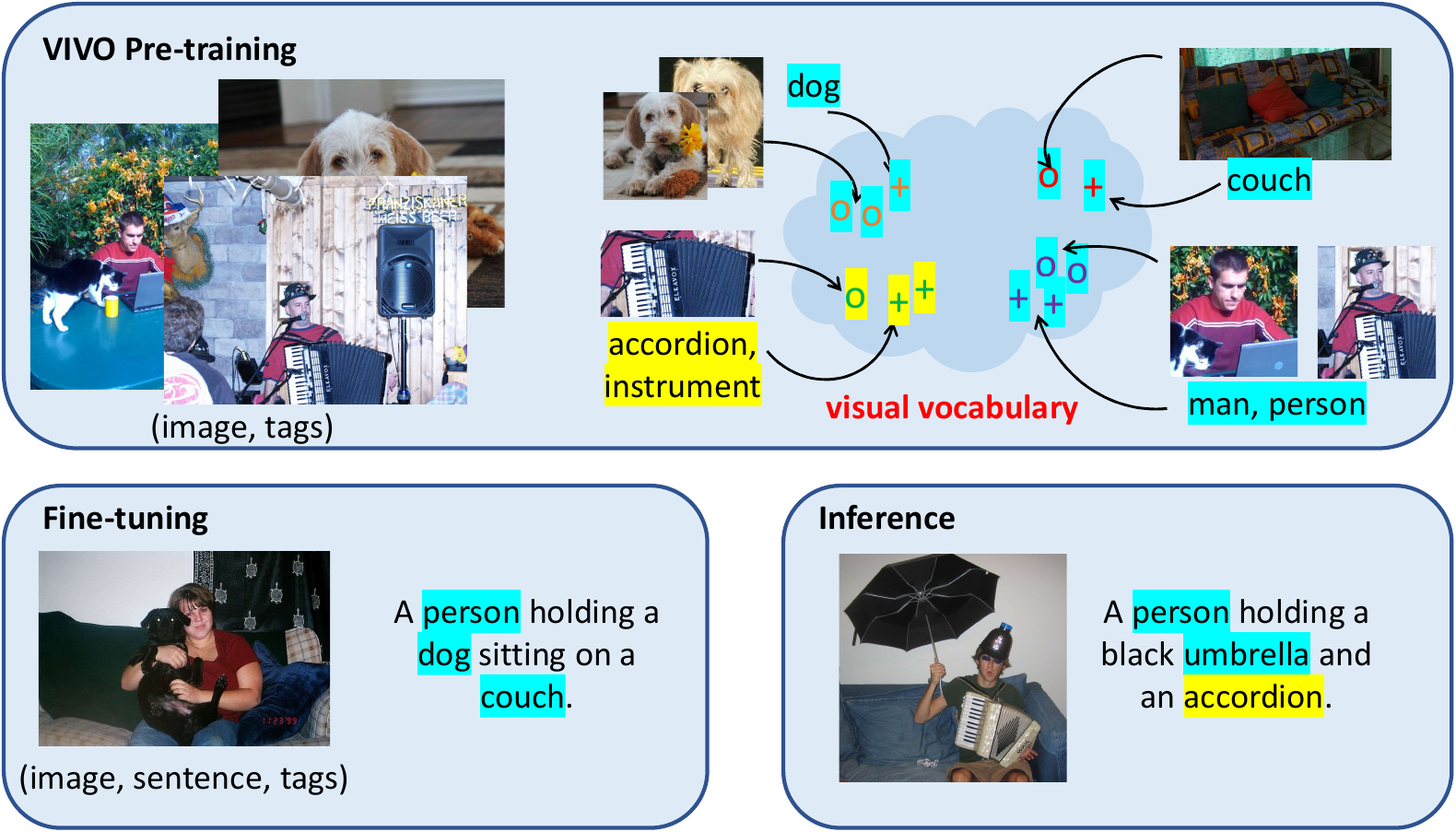}
\vspace{-2mm}
\figcaption{VIVO pre-training uses paired image-tag data to learn a rich visual vocabulary where image region features and tags of 
% the same object are aligned.
the semantically similar objects are mapped into vectors that are close to each other.
Fine-tuning is conducted on paired image-caption data that only cover a limited numbers of objects (in blue). During inference, our model can generalize to describe novel objects (in yellow) that are learnt during VIVO pre-training.}
\label{fig:concept}
% \end{center}
\end{figure}

Image captioning is a long-standing task in artificial intelligence~\cite{farhadi2010every, kulkarni2013babytalk, kuznetsova2012collective, mitchell2012midge, yang2011corpus, fang2015captions}. The task is challenging in  
% complex task 
that it requires visual perception and recognition, 
% real-world knowledge in language space, 
and natural language generation grounded in perception and real-world knowledge ~\cite{kuznetsova2012collective, yang2011corpus}. With recent progress in % the fields of 
computer vision~\cite{he2017mask, ren2015faster}, natural language processing~\cite{devlin2018bert,Radford2018ImprovingLU, vaswani2017attention}, and vision-language understanding~\cite{li2020oscar, sharma2018conceptual, zhou2020unified}, the performance on image captioning has been substantially improved on public benchmarks like COCO~\cite{chen2015microsoft} and Flickr30k~\cite{young2014image}. However, models trained on such datasets with limited visual concepts generalize poorly to in-the-wild images~\cite{tran2016rich}.

To improve image captioning in the wild, the nocaps benchmark~\cite{agrawal2019nocaps} is developed to evaluate Novel Object Captioning (NOC)\footnote{We use ``NOC'' to represent the task of novel object captioning and ``nocaps'' to refer to the nocaps benchmark.} at scale.
The training data for nocaps is the COCO dataset consisting of image-caption pairs and the Open Images dataset~\cite{kuznetsova2018open} containing bounding boxes and image-level tags. %labels. 
The test data consists of images selected from Open Images, containing nearly $400$ objects that are not or rarely seen in the COCO dataset. 
This raises the challenge of how to generate captions that describe novel objects unseen in the paired image-caption training data. 
A common strategy is to resort to alternative data sources without caption supervision. % for NOC. 
%The test set consists of nearly $400$ novel objects from Open Images that are unseen 
%not present
%in the caption corpora of COCO. 
Prior works on NOC~\cite{lu2018neural, wu2018decoupled} propose to generate template sentences that can be filled in with detected visual concepts for NOC.
However, the relationship between image and text is not fully explored in their frameworks. 
We will show that the performance of NOC can be significantly improved by pursuing image-text aligned representation learning.

In this paper, 
we present VIsual VOcabulary (VIVO) pre-training that leverages large amounts of vision data without caption annotations to learn a rich visual vocabulary for NOC. As shown in Figure~\ref{fig:concept}, we define visual vocabulary as a joint embedding space where image region features and tags of 
% the same object or 
 semantically similar objects are mapped into vectors that are close to each other, \eg, ``person'' and ``man'', ``accordion'' and ``instrument''. Once the visual vocabulary is pre-trained, we can fine-tune the model using image-caption pairs for caption generation. 
Note that the dataset used for fine-tuning only covers a small subset of the most commonly occurred objects in the learnt visual vocabulary. 
Nevertheless, our model can generalize to any images that contain similar scenes (\eg, people sitting in couch in Figure~\ref{fig:concept}) with novel objects unseen in the fine-tuning dataset, like ``accordion'', thanks to the pre-trained visual vocabulary.
% learnt in the pre-training stage.}

The VIVO pre-training method is motivated to learn the cross-modality semantic alignment, similarly as in
%shares the same motivation as 
 conventional Vision-Language Pre-training (VLP) methods.
%to learn the cross-modality semantic alignment. 
However, unlike existing VLP models which are pre-trained using image-caption pairs, VIVO is pre-trained on image-tag pairs. 
To the best of our knowledge, VIVO is the first VLP method that does not rely on caption annotations. Thus, it opens the possibility of leveraging, for VLP, many existing vision datasets originally developed for image tagging or object detection tasks like ImageNet~\cite{deng2009imagenet}, Open Images~\cite{kuznetsova2018open}, Objects365~\cite{shao2019objects365}, etc. 
% Moreover, we can also use machine-generated tags, as weak supervision signals,
% thus making it possible 
% to leverage large amounts of unlabeled images for VLP.
Moreover, we can also leverage large amounts of images, paired with machine-generated tags as weak supervision signals, for VLP.
% in a large scale.}

VIVO pre-training aims to learn a joint representation of visual and text input.  
%is conducted 
%at the sub-sentence level 
We feed to a multi-layer Transformer model an input consisting of image region features and a paired image-tag set.
We then randomly mask one or more tags, and ask the model to predict these masked tags conditioned on the image region features and the other tags.
%as input modalities. 
% We construct a sequence from a set of tags as input for the Transformer model. 
Given that tags are not ordered, we employ the Hungarian matching loss~\cite{stewart2016end, carion2020end} for tag prediction optimization. % in pre-training,
% which effectively encourages 
% aiming to learn a joint representation for both visual and text. 
Extensive experiments show that VIVO pre-training significantly improves the captioning performance on NOC. In addition, our model can precisely align the object mentions in a generated caption with the regions in the corresponding image.
% ground the generated object mentions in image regions.}

In summary, we make the following contributions.
\begin{tight_itemize}
\item{We propose a new VIVO pre-training method that leverages large amounts of vision data without caption annotations for vision-language representation learning.}
%\item We propose VIVO pre-training that leverages vision datasets without captions to learn a rich visual vocabulary.
\item{We develop a Hungarian matching loss with masked tag prediction to conduct pre-training with image-tag pairs.}
%\item We uncover the secret behind VIVO pre-training by analyzing visual-text alignment quantitatively.
%\item We interpret the benefit of VIVO pre-training by analyzing visual-text alignment quantitatively.
\item With a single model, our method achieves the new state-of-the-art result on the nocaps benchmark and surpasses the human CIDEr score. 
%for the first time. %With a single model, our method surpasses human performance in novel object captioning on nocaps benchmark.  
\end{tight_itemize}

\begin{figure*}[t]
\begin{center}
\includegraphics[width=0.9\textwidth]{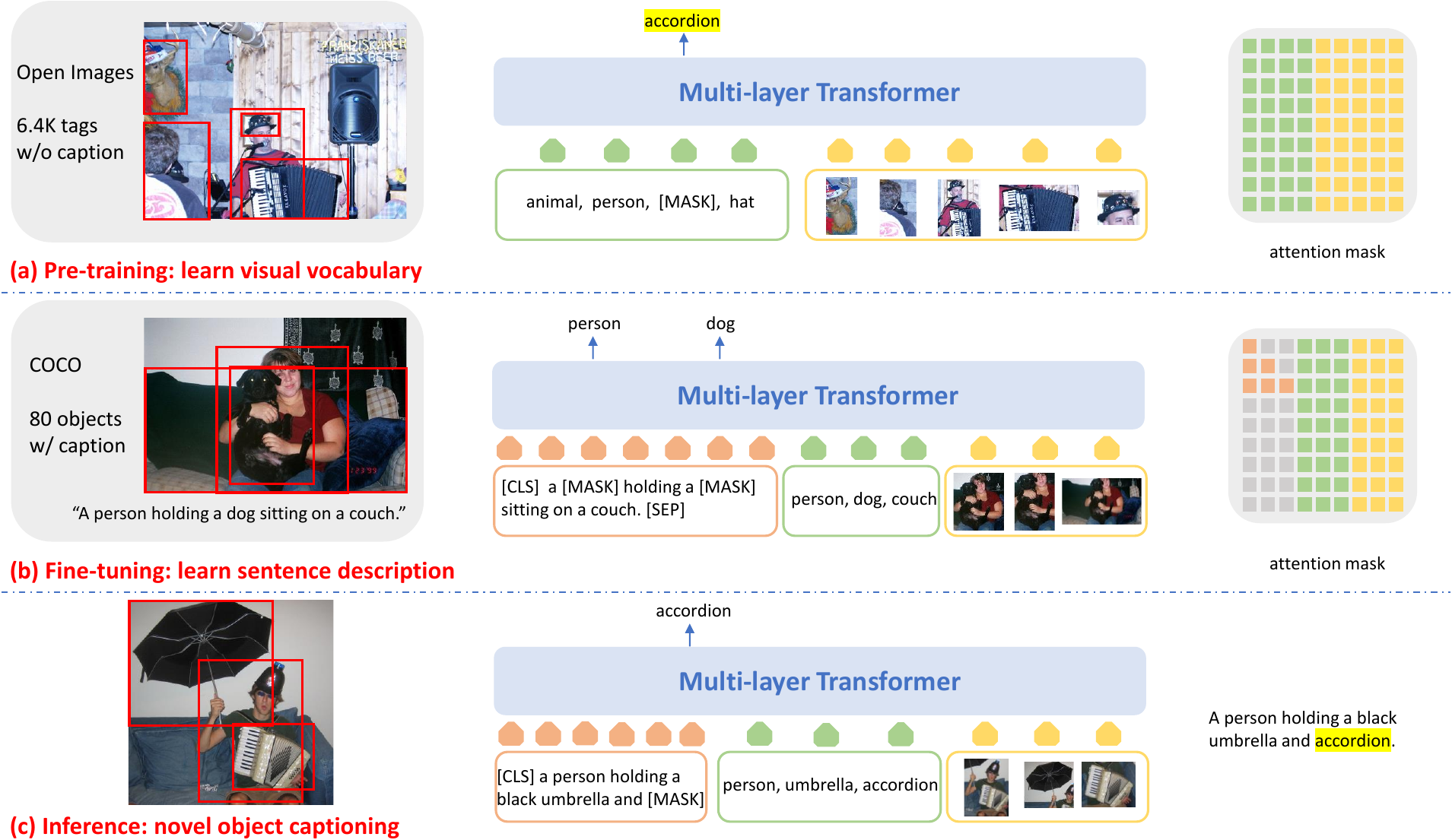}
\figcaption{
%The proposed disentangled training scheme. (a) In pre-training, we leverage visual tags as anchor points to align visual regions and thus learn a rich visual vocabulary. (b) During the fine-tuning stage, we use image-sentence paired data to learn how to describe objects in the image. (c) During inference, our model can generate captions capable of describing novel objects.
The proposed two-stage training scheme. (a) In VIVO pre-training, we train a Transformer-based model on image-tag pairs for tag prediction, where it learns cross-modal representations for rich visual concepts. (b) In fine-tuning, we train the same model on limited image-caption pairs to learn how to generate captions conditional on the image and tags. (c) During inference, given the image and detected tags, our model is applied iteratively to generate a sequence of words describing novel objects in an auto-regressive manner.
}
\label{fig:overview}
\end{center}
\end{figure*}

\section{Prior Work}
\Paragraph{Image Captioning} \xiyin{Prior works on image captioning have focused on % various aspects including 
exploring different model structures and learning methods for different applications. 
For example, ~\citet{song2019connecting, wang2019hierarchical, gao2019deliberate, huang2019attention, pan2020x, guo2020normalized, cornia2020meshed} explore different attention mechanisms in captioning modeling. Other works improve the performance with reinforcement learning~\cite{rennie2017self, li2019meta, yang2020fashion} or adversarial learning~\cite{chen2019improving, dognin2019adversarial}. 
Different applications such as dense captioning~\cite{johnson2016densecap,yin2019context, li2019learning}, grounded captioning~\cite{ma2019learning, zhou2020more}, image captioning with reading comprehension~\cite{sidorov2020textcaps} have been studied. 
However, all these methods assume that most of the visual objects in test data are seen in training data. 
% consistency between training and testing data 
Thus, they do not work well for NOC, where the objects presented in test images are often unseen in the caption-annotated training data.}

%\xiyin{Unsupervised image captioning, where no image-sentence pairs are used, shares some similarities to our work. It is first proposed in~\cite{feng2019unsupervised} and tackled with knowledge distillation and an adversarial objective.~\cite{laina2019towards} formulates a cross-modality alignment via visual concepts to solve this problem. Unsupervised image captioning is a more challenging task than NOC and is not the focus of our work.}

\Paragraph{Novel Object Captioning (NOC)} NOC requires a model to generate image captions that describe novel objects that are unseen in the paired image-caption training data. 
% It is a challenging task that has been explored in recent years. 
Since the task setting resembles that in real-world applications, it draws growing interest in the research community.
The early works, such as Deep Compositional Captioner~\cite{hendricks2016deep} and Novel Object Captioner~\cite{venugopalan2017captioning}, propose to use unpaired image and sentence data to transfer knowledge among semantically similar visual concepts. Empirical evaluation on the COCO dataset by holding out $8$ novel object categories suggests that these methods might be applicable to NOC. % for NOC at a small scale. 

Recent studies propose to explicitly leverage the object detection results for NOC. \citet{yao2017incorporating} use LSTM-C with a copying mechanism to assemble the detected novel objects for caption generation. 
Neural Baby Talk~\cite{lu2018neural} and Decoupled Novel Object Captioner~\cite{wu2018decoupled} generate template sentences that are later filled in with visual concepts recognized by object detectors. 
Similarly, Constrained Beam Search~\cite{anderson2016guided} is exploited to generate captions that contain detected novel objects~\cite{agrawal2019nocaps}. 

None of the aforementioned methods for NOC fully exploits the relationship between image and text, which we argue is crucial to the quality of generated captions. 
% Though the above methods work for NOC, the relationship between image and text is not fully explored. 
In this study, we pre-train a Transformer model to learn a visual vocabulary where object tags are aligned with their corresponding image feature representations in a semantic space.
% with those of tags to improve NOC via VIVO pre-training. 

\Paragraph{Vision and Language Pre-training} 
%Large-scale VLP on paired image-caption data % to learn cross-modal representations is shown effective to improve many vision-language tasks.
%VLP on a large-scale of image-text paired training data is shown to be effective for learning the cross-modal representations for the VL downstream tasks. 
Motivated by BERT~\cite{devlin2018bert}, many VLP methods have been proposed to learn vision-language representations by pre-training large-scale Transformer models ~\cite{lu2019vilbert,tan2019lxmert,su2019vl,chen2020uniter, zhou2020unified,li2020oscar}.
%Specifically, previous work has focused on different model architectures including single-stream~\cite{chen2020uniter, su2019vl, zhou2020unified} and two-stream approaches~\cite{lu2019vilbert,tan2019lxmert}, the loss functions like masked modality modeling~\cite{li2020unicoder, huang2020pixel} and matching loss~\cite{li2020oscar, chen2020uniter}, and various downstream tasks. 
Most existing VLP methods are developed for understanding tasks such as image-text retrieval and visual question answering.
Only a few of them ~\cite{zhou2020unified, li2020oscar} can be applied to image captioning. 
But these methods use paired image-caption data for pre-training, and are not applicable to NOC.
% When it comes to NOC, where no additional paired image-caption training data is available, none of the above VLP methods can be applied.
In this study, we break the dependency on image-caption pairs in VLP for the first time. The proposed VIVO pre-training learns vision-language alignment on image-tag pairs, improving the image captioning results on both NOC and the general image captioning task.
%at the sub-sentence level by leveraging image-level tags for cross-modality alignment. 
%The effectiveness of VIVO is demonstrated on both NOC and the general image captioning task. 

\section{Proposed Method}

Recent image captioning models have achieved impressive results on the tasks where large amounts of paired image-caption training data is available. 
But they generalize poorly to images in the wild, where there are a wide variety of visual objects that are unseen in the caption corpora for training. 
For example, the models trained on COCO Captions can faithfully describe images containing objects such as ``people'', ``dogs'', or ``a couch'', but fail to generate a reasonable caption for any image containing ``an accordion'' since the object is unseen in COCO Captions. 

To address this problem, we propose a weakly supervised learning approach to pre-training image captioning models on image-tag pairs that, compared to image-caption pairs, are of larger amounts and contain many more diverse visual objects.
%To enable the model to learn from image-tag pairs to generate captions, we 
Our approach uses a two-stage training scheme that consists of VIVO pre-training and fine-tuning.
%First, we pre-train a Transformer model on large amounts of image-tag pairs and ask the model to predict missing tags given a bag of tags and image regions. Then, the pre-trained Transformer model is fine-tuned on a small dataset with caption annotations. Given image regions and tags, the model learns to predict the conditional caption sentence where some positions are randomly masked out. To make the model generate sentence from left to right at inference time, during fine-tuning we also apply the uni-directional attention mask on caption sequence to prevent the positions from attending to subsequent positions. For inference, given the image and detected tags, the fine-tuned model will be applied iteratively to generate the natural language tokens one by one.
%To illustrate how the model is learned to generate a  caption of an image that includes novel objects which are unseen in image-caption data during training, consider the example in Figure~\ref{fig:overview}.
Figure~\ref{fig:overview} illustrates our approach using an example.
First, in the pre-training stage (Figure~\ref{fig:overview}(a)), an image captioning model learns to label image regions using tags (\eg, ``person'', ``accordion'') using image-tag pairs as training data, where the object ``accordion'' is included. 
Then in fine-tuning (Figure~\ref{fig:overview}(b)), given image-caption pairs and their corresponding object tags detected (\eg, ``person'' and ``dog''),
% the model learns to map an image to a (reusable) caption template (\eg, ``[A] holding [B] ...''), and fill the template with the object tags to form a caption (\eg, ``a person holding a dog.''). While the caption templates are learned from image-caption pairs, the object tags to be filled may refer to novel visual objects that are unseen in image-caption pairs (but seen in image-tag data in this example).
the model learns to map an image to a sentence conditioned on the detected objects, \eg, ``[A] holding [B] ...'', where [A] and [B] could attend to object tags. While the sentences are learned from image-caption pairs, the object tags may refer to novel visual objects that are unseen in image-caption pairs (but seen in image-tag data in this example).
Thus, our model achieves the compositionality generalization, allowing for zero-shot generalization to novel objects for image captioning.
%Although the model is not exposed to the captions containing the novel object mention ``accordion'', it learns how to select a proper template, and fill it with object mentions to compose a caption.   
% it stills learns the constituent expressions from available tags ``person'', ``dog'', and the rules to combine them. 
As shown in Figure~\ref{fig:overview}(c), at inference time the model is able to
% select the template ``[A] holding [B] ...'', fill it with the object tags ``person'' and ``accordin'', which are unseen in the paired image-caption training data, and compose the caption ``a person holding an accordion''.
recognize objects (\eg, ``person'', ``accordion'') and compose familiar constituents in a novel way to form a caption ``a person holding an accordion''. 
%make a novel composition of familiar constituents, e.g., the object ``accordion'' learned from pre-training and the template ``[A] holding [B] ...'' learned from fine-tuning, to form a sentence as ``person holding accordion ...''. 

The model architecture is shown in Figure~\ref{fig:model_arch}. 
It consists of multiple Transformer layers to encode the input into a feature vector and a linear layer with softmax to generate the text description of the visual objects in the image. % the learned visual vocabulary.
In what follows, we describe in detail the way the model is pre-trained and fine-tuned.

\subsection{VIVO Pre-training}
We pre-train the Transformer model on a large-scale dataset with abundant tags, \eg, the Open Images training set with $6.4K$ classes of image-level tags.
Unlike many existing VLP methods that rely on image-caption pairs, VIVO pre-training is conducted solely on image-tag pairs, which are much easier to collect by either human labeling or auto tagging.
The training objective is to predict the missing (masked) tags given a bag of image-level tags and image regions.
We denote the training set as $\mathbb{D} = \{{\bf{I}}_i, {\bf{G}}_i\}_{i=1}^{N}$ with $N$ images and their corresponding tags, where ${\bf{G}}_i = \{g_{ij}\}_{j=1}^{L_i}$ is a set of $L_i$ image-level tags that are associated with the image ${\bf{I}}_i$. 
These tags are textual labels of the visual objects presented in the image, \eg, ``person'', ``cat'', ``dinning table'', etc.
%They are from either human annotations and/or pseudo labels generated from a tagging or a detection model.
In the rest of the paper, we omit the subscript $i$ for simplicity. 

%We use a multi-layer Transformer as our model structure, because it has shown great performance in learning cross-modality representations in vision-language tasks~\cite{chen2020uniter, li2020oscar}. We use a multi-layer Transformer to learn a joint representation for both vision and language domains. The input to the Transformer consists of two parts. The first part is visual input of image region features ${\bf{V}} = \{{\bf{v}}_k\}_{k=1}^{K}$, which are extracted from image $\bf{I}$ using a Faster R-CNN object detector trained on Visual Genome dataset~\cite{anderson2018bottom}. The second part is textual input of tag tokens ${\bf{T}} = \{t_j\}_{j=1}^{T}$, which are tokenized from the tags in $\bf{G}$. During training, some tokens are randomly masked out for the model to predict. In order to predict the masked tag, the model will have to resolve to other information like image region features and other tags. Therefore, the tag prediction objective encourages the cross-modality alignment between image regions and tags.

We use a multi-layer Transformer model to learn a joint representation for both vision and language domains. The input to the Transformer model consists of image region features ${\bf{V}}$ and tag tokens ${\bf{T}}$, where ${\bf{V}} = \{{\bf{v}}_k\}_{k=1}^{K}$ are extracted from image $\bf{I}$ using a detector trained on Visual Genome dataset~\cite{anderson2018bottom}, and ${\bf{T}} = \{t_j\}_{j=1}^{T}$ are tokenized tags in $\bf{G}$. During training, some tokens are randomly masked out for the model to predict.
%In order to predict the missing tag, the model will have to resolve to image region features and other tags. Therefore, the tag prediction objective encourages to learn a cross-modality alignment between image regions and tags.

%The goal of VIVO pre-training is to learn a joint representation for both vision and language domains, where the final representation output of the image regions will be aligned with the corresponding visual concepts in the word vocabulary - we term such subset of vocabulary as visual vocabulary (Figure~\ref{fig:concept}). Similar to other VLP work, we apply mask language modeling on our tags for masked tag prediction, as shown in Figure~\ref{fig:overview} (a). The tokens are randomly masked out with $15\%$ probability. In order to predict the masked tag, it will have to resolve to other information like image region features. Therefore, such a pre-training objective encourages the cross-modality alignment between image regions and tags. 

The main difference between a caption and a set of tags is that words in the caption are ordered while tags are not ordered. This unordered nature may result in ambiguity in tag prediction when two tags are masked out simultaneously. For example, if the masked tokens are ``dog" and ``cat", we can predict each token in either position without restricting to the original position or order in the input. To resolve this issue, we propose to use the Hungarian matching loss~\cite{stewart2016end, carion2020end} to formulate the tag prediction as a set-matching problem. 

We denote the set of $M$ masked tokens as $\tilde{{\bf{T}}} = \{t_m\}_{m=1}^{M}$ where $t_m$ is the token id in the vocabulary, and the prediction probabilities of the corresponding representations in the final layer of Transformer as ${\bf{P}} = \{{\bf{p}}_i\}_{i=1}^{M}$ where ${\bf{p}}_i$ is the classification probabilities for the $i$-th masked position. Since the target tokens in $\tilde{{\bf{T}}}$ are unordered, we need an one-to-one mapping from $\tilde{{\bf{T}}}$ to ${\bf{P}}$ such that the prediction for each masked position is assigned one of the target tokens. Once such an assignment $\alpha$ is known, the loss is defined as:
\begin{equation}
\label{eq:loss}
L(\tilde{{\bf{T}}}, {\bf{P}}, \alpha) = \sum_{i=1}^{M} (- \log({\bf{p}}_{i}(t_{\alpha(i)}))) 
\end{equation}
where $\alpha$ is a permutation of the $M$ indices, i.e., $\alpha(i)$ is the index of the target token assigned to the $i$-th prediction. Since the assignment is unknown, we want $\alpha$ to be the best possible mapping between $\tilde{{\bf{T}}}$ and ${\bf{P}}$. Formally, we define such best possible $\alpha$ to be the one that minimizes the following total cost among all the valid\footnote{For a tag tokenized into multiple tokens, the order of tokens within the tag cannot be changed.} permutations:
\begin{equation}
\label{eq:alpha}
\hat{\alpha} = \underset{\alpha}{\argmin}\sum_{i=1}^{M}C( {\bf{p}}_{i}, t_{\alpha(i)}),
\end{equation}
where $C({\bf{p}}_i, t_m) = 1 - {\bf{p}}_i(t_m)$ is the cost function of assigning the target $t_m$ to the $i$-th prediction. The reason why we use $C({\bf{p}}_i, t_m)$ instead of $-\log({\bf{p}}_{i}(t_{\alpha(i)}))$ as in \eqref{eq:loss} is that it is bounded. Now we can compute the final loss as $L(\tilde{{\bf{T}}}, {\bf{P}}, \hat{\alpha})$, where $L$ is defined in \eqref{eq:loss} and $\hat{\alpha}$ is defined in \eqref{eq:alpha}.

As shown in Figure~\ref{fig:overview} (a), we use bi-directional attention mask in VIVO pre-training.
In order to predict a missing tag, the model will have to resort to image region features and the other tags. So it learns a joint representation containing information from both image regions and textual tags. This facilitates the cross-modality alignment between representations of image regions and tags.
%After VIVO pre-training, we expect the model to be capable in recognizing the trained visual concepts in different images.

\iffalse
The task is to assign the $M$ token ids as the targets for $M$ predictions such that the loss is minimized. To achieve this, we first search a permutation of $M$ indexes $\alpha$ where $\alpha(i)$ represents the index of the targeted token id of the $i^{th}$ masked token. The optimal assignment is computed as:
\begin{equation}
\hat{\alpha} = \underset{\alpha}{\argmin}\sum_{i}^{M}L(t_i, {\bf{p}}_{\alpha(i)}),
\end{equation}
where $L(t_i, {\bf{p}}_{\alpha(i)})$ is the loss by assigning ground-truth token id $t_i$ to the prediction of ${\bf{p}}_{\alpha(i)}$. We use standard softmax cross-entropy loss for classification. The final loss is calculated with the optimal assignment:
\begin{equation}
L({\bf{T}}_m, {\bf{P}}) = \sum_{i}^{M} (- \log({\bf{p}}_{\alpha(i)}(t_i))) 
\end{equation}

As shown in Figure~\ref{fig:overview} (a), we use bi-directional attention mask to facilitate the cross-modality alignment between image regions and tags via the Hungarian matching loss in masked tag prediction. After VIVO pre-training, we expect the model to be capable of recognizing the trained visual concepts in different images.
\fi

\subsection{Fine-tuning and Inference}
After pre-training, the Transformer model is fine-tuned on a dataset where both captions and tags are available, \eg, the COCO set annotated with tags from $80$ object classes and captions. The tags can also be automatically generated using a pre-trained tagging or detection model. Given image regions and tags, the model learns to predict the conditional caption sentence where some positions are randomly masked out. More specifically, the input to the model during fine-tuning is a triplet of image region features ${\bf{V}}$, a set of tags $\bf{T}$ and a caption ${\bf{C}}$, where ${\bf{V}}$ and $\bf{T}$ are constructed in the same way as described in pre-training, and ${\bf{C}}$ is a sequence of tokens.
% by tokenizing the caption sentence. 
During fine-tuning, we randomly mask out some of the tokens in a caption sentence for prediction, and optimize the model parameters using the cross-entropy loss. To make the model generate captions from left to right at inference time, during fine-tuning we apply the uni-directional attention mask on a caption sequence to prevent the positions from attending to subsequent positions.

During inference, we first extract image region features and detect tags from a given image. Then the model is  applied to generate a sequence, one token at a time, until it outputs the end of sentence token or reaches the maximum length. At each step the model is auto-regressive, consuming the previously generated tokens as additional input when generating the next.

In the next section, we present extensive experimental results, showing that our model can generate captions to describe novel objects and 
that the alignment between image regions and tags, learned from VIVO pre-training, is crucial to the model's superior performance on NOC.

\section{Experiments}
\subsection{Experimental Settings}
\Paragraph{Datasets} We use the Open Images V5 challenge training set, which has $1.7M$ images, for VIVO pre-training. We select $500$ classes\footnote{Only $500$ out of $600$ objects are used in the challenge set, as we further refine the labels by removing classes that are ``parts'' (\eg, human eyes).} from bounding box annotations and $6.4K$ classes from human verified image-level labels. 
%The object labels and image-level labels are used jointly for VIVO pre-training.
The joint image-tag pairs, containing $6.4K$ unique classes in total, are used in VIVO pre-training.
In the fine-tuning stage, the training data is the COCO training set of $118K$ images, each with $5$ captions. 
We evaluate our model on the validation and test sets of nocaps, which consist of $4.5K$ and $10.6K$ images from the Open Images validation and test sets, respectively. 

\Paragraph{Implementation Details} We use the object detector from UpDown~\cite{anderson2018bottom} to extract image region features, \xiyin{which are concatenated with scaled bounding boxes to form a $2054$-dimension vector \leizhang{($2048$D for the visual features and $6$D for the bounding box encoding including top-left and bottom-up corners as well as the box's width and height)}. 
We use an object detector trained on the Open Images dataset to detect object tags for all datasets. For pre-training and fine-tuning, we also add the ground-truth tags from the training sets.}
%We use an object detector trained on Open Images to extract labels for fine-tuning and inference. We also add the ground-truth labels from COCO training set. 
No ground-truth tags are used on the nocaps validation and test sets.
The Transformer model is initialized using BERT-base~\cite{devlin2018bert} where we add a linear layer to transform the image region features to the vectors with same size as the word embeddings. 

In VIVO pre-training, we use a maximum of $50$ image regions and $15$ tag tokens per image. The model is trained for $160K$ iterations (about $100$ epochs) with a batch size of $1024$ and a learning rate of $5\times10^{-5}$. 
In fine-tuning, we set the maximum caption length to $40$ and the maximum tag length to $30$. The model is trained for $30$ epochs with a batch size of $256$ and a learning rate of $5\times10^{-5}$, optimized using the cross-entropy loss. To further boost the performance, we perform the SCST optimization~\cite{rennie2017self} with a learning rate of $2\times10^{-6}$ for $5$ epochs. During inference, we use greedy decoding to generate image captions with a maximum length of $20$.

\subsection{Novel Object Captioning}
We compare our method with UpDown~\cite{anderson2018bottom, agrawal2019nocaps} and OSCAR\footnote{We compare with OSCAR base whose model size is the same as ours. In fact, our model with $12$ layers and hidden size of $768$ even outperforms the OSCAR large model.}~\cite{li2020oscar}, which holds the state-of-the-art result on the nocaps benchmark. The training data for the baselines is the COCO dataset. Following prior settings, we also report the results after our model is optimized using SCST~\cite{rennie2017self} and generates captions using Constrained Beam Search (CBS)~\cite{anderson2016guided}.
%In this subsection, we conduct VIVO pre-training on the training set of Open Images and evaluate on the nocaps benchmark. we compare our method with the popular UpDown approach~\cite{anderson2018bottom, agrawal2019nocaps} which is trained on COCO Captions with image region features and caption annotations, and OSCAR~\cite{li2020oscar} which uses object tags as an additional modality to train image captioning and achieved the state-of-the-art performance on nocaps. Following prior works, we also add Constrained Beam Search (CBS)~\cite{anderson2016guided} during inference to enforce the model to generate captions including desired objects. We use predicted object tags with high precision as our constraints. 

The evaluation results on nocaps \kevin{validation} and test sets are shown in Table~\ref{tab:nocaps_caption}. By leveraging VIVO pre-training on the Open Images dataset, our method has achieved significant improvement compared to all prior works. Our plain version (VIVO) already outperforms UpDown+ELMo+CBS and OSCAR by a large margin. 
It is worth noting that CBS brings absolute gains of $17.8\%$ and $15.5\%$ for UpDown and OSCAR, respectively, but it only improves VIVO by $3.8\%$. This suggests that our model is more capable of generating captions with novel objects without explicitly adding any constrains.
%\xiaowh{We also observe that, as shown in the first three images of Figure~\ref{fig:nocaps}, despite of the benefit of deterministic mentions of a given object brought from CBS, the model sometimes generates broken sentences due to the low probability of the CBS-enforced word, which breaks the statistics learned in the language model. Our VIVO model can alleviate this problem by predicting the words of novel objects with higher probabilities.}
Our best results are new state-of-the-art and surpasses the human CIDEr score on the overall dataset.

\begin{table*}[ht]
\centering
\small
\begin{tabular}{l@{\hspace{2mm}}|c@{\hspace{2mm}}c@{\hspace{2mm}}|c@{\hspace{2mm}}c@{\hspace{2mm}}|c@{\hspace{2mm}}c@{\hspace{2mm}}|c@{\hspace{2mm}}c}
\toprule
\multirow{2}{*}{method } & \multicolumn{2}{c|}{in-domain} & \multicolumn{2}{c|}{near-domain} & \multicolumn{2}{c|}{out-of-domain} & \multicolumn{2}{c}{overall}\\ 
& CIDEr & SPICE & CIDEr & SPICE & CIDEr & SPICE & CIDEr & SPICE \\ \midrule
\multicolumn{9}{c}{Validation Set} \\ \hline
UpDown~\cite{agrawal2019nocaps} & $78.1$ & $11.6$ & $57.7$ & $10.3$ & $31.3$ & $8.3$ & $55.3$ & $10.1$ \\
UpDown + CBS & $80.0$ & $12.0$ & $73.6$ & $11.3$ & $66.4$ & $9.7$ & $73.1$ & $11.1$ \\
UpDown + ELMo + CBS & $79.3$ &  $12.4$ & $73.8$ & $11.4$ & $71.7$ & $9.9$ & $74.3$ & $11.2$ \\
\midrule
\textsc{Oscar}~\cite{li2020oscar} & $79.6$ & $12.3$ & $66.1$ & $11.5$ & $45.3$ & $9.7$ & $63.8$ & $11.2$ \\
\textsc{Oscar} + CBS & $80.0$ & $12.1$ & $80.4$ & $12.2$ & $75.3$ & $10.6$ & $79.3$ & $11.9$ \\
\textsc{Oscar} + SCST + CBS & $83.4$ & $12.0$ & $81.6$ & $12.0$ & $77.6$ & $10.6$ & $81.1$ & $11.7$ \\
\midrule
\rowcolor{Gray} 
% VIVO & $85.9$ & $12.8$ & $78.1$ & $12.2$ & $67.8$ & $10.4$ & $77.1$ & $11.9$ \\
VIVO & $88.8$ & $12.9$ & $83.2$ & $12.6$ & $71.1$ & $10.6$ & $81.5$ & $12.2$ \\
\rowcolor{Gray} 
% VIVO + CBS & $85.6$ & $12.5$ & $81.6$ & $12.0$ & $80.6$ & $10.5$ & $81.9$ & $11.8$ \\
VIVO + CBS & $90.4$ & $13.0$ & $84.9$ & $12.5$ & $83.0$ & $10.7$ & $85.3$ & $12.2$ \\
\rowcolor{Gray} 
VIVO + SCST + CBS & $\bf 92.2$ & $12.9$ & $\bf 87.8$ & $12.6$ & $87.5$ & $11.5$ & $\bf 88.3$ & $12.4$ \\
\midrule
Human             & $84.4$ & $\bf 14.3$ & $85.0$ & $\bf 14.3$ & $\bf 95.7$ & $\bf 14.0$ & $87.1$ & $\bf 14.2$ \\
\bottomrule
\multicolumn{9}{c}{Test Set} \\ \hline
\rowcolor{Gray}
VIVO + SCST + CBS & $\bf 89.0$ & $12.9$ & $\bf 87.8$ & $12.6$ & $80.1$ & $11.1$ & $\bf 86.6$ & $12.4$ \\
Human & $80.6$ & $\bf 15.0$ & $84.6$ & $\bf 14.7$ & $\bf 91.6$ & $\bf 14.2$ & $85.3$ & $\bf 14.6$ \\
\bottomrule
\end{tabular}
\vspace{-2mm}
\caption{Evaluation on nocaps validation and test sets.}
\label{tab:nocaps_caption}
\tabvspace
\end{table*}

%Given the fact that it is possible to achieve good BLEU and CIDEr scores without mentioning the novel object, to further evaluate the model's ability to integrate novel objects in the captions,   
% Kevin: it may not be good to challenge CIDEr and BLEU metrics here. We actually use CIDEr as our golden evaluation metric when comparing with human CIDEr. Maybe we just simply describe our goal here.
\kevin{To quantitatively evaluate how well the model can describe novel objects,} 
we also calculate the F1-score following~\citet{hendricks2016deep}, where all the objects mentioned in the generated caption sentences are compared against the ground-truth object tags. Table~\ref{tab:nocaps_f1} shows the comparison with OSCAR on the nocaps validation set.
%\xiaowh{In nocaps evaluation, the main difference between VIVO and OSCAR is that the VIVO model is pre-trained on Open Images while OSCAR is initialized from BERT. When fine-tuning on COCO Captions, OSCAR and VIVO use the same input, \ie, image region features, captions, and object tags.} % Kevin: I feel this part could still confuse the readers if they are not familiar with nocaps's rule. Maybe we could put these configuration stuff at the beginning of this section.  
%We search to tune hyperparameters for both OSCAR and VIVO when fine-tuning on COCO and evaluate under the best settings. 
%It is obvious that VIVO pre-training improves the F1 scores substantially especially for out-of-domain objects.
% Compared with OSCAR, 
We see that VIVO improves OSCAR in F1-scores substantially especially for out-of-domain objects.
This again verifies the effectiveness of VIVO pre-training in learning to recognize novel objects for NOC.

Although object tags are used in both VIVO pre-training and fine-tuning stages, we show that 
%the success of mentioning novel objects at inference time 
the model's capability of generating captions that precisely describe novel objects at inference time
attributes largely to
%relies on 
pre-training. 
We compare the distribution of object tags on COCO and nocaps, which are generated by the object detector trained on the Open Images dataset and used for fine-tuning and inference, respectively. 
As shown in Table~\ref{tab:nocaps_dist}, COCO has a long-tail distribution where $415$ out of $568$ categories amounts only to $2.43\%$ of all the tags. 
The under-representation of novel objects makes the trained model statistically unlikely to generate 
plausible captions that describe these novel objects. 
Therefore, our VIVO pre-training, which 
% alleviates 
mitigates 
the data imbalance issue by leveraging diverse tags in image-tag pairs, is crucial to improving model's generalization property, as empirically demonstrated on NOC.

\begin{table}[h]
\centering
\small
\begin{tabular}{l@{\hspace{2mm}}|c@{\hspace{2mm}}c@{\hspace{2mm}}c}
\toprule
model  & in-domain & out-of-domain & entire  \\ \midrule
OSCAR~\cite{li2020oscar} & $39.5$ & $15.7$ & $20.7$ \\
VIVO & $\bf 46.3$ & $\bf 30.6$ & $\bf 33.8$ \\ 
\bottomrule
\end{tabular}
\vspace{-2mm}
\caption{Comparison of F1-scores (in \%) on object classes of Open Images, evaluated on the nocaps validation set. There are $504$ classes in total. $105$ of them are in-domain, which are $80$ common classes from COCO and $25$ objects frequently appearing in COCO Captions. The remaining $399$ classes are the out-of-domain objects.}
\label{tab:nocaps_f1}
\end{table}

\begin{table}[h]
\centering
\small
\begin{tabular}{l@{\hspace{2mm}}|c@{\hspace{2mm}}c@{\hspace{2mm}}c@{\hspace{2mm}}c@{\hspace{2mm}}c}
\toprule
$\#$occur in COCO ($<=$) & $0$ & $10$ & $100$ & $1K$ & $10K$ \\ \midrule
$\#$categories & $194$ & $274$ & $415$ & $522$ & $563$ \\
percentage in COCO  &$0.0$ & $0.14$ & $2.43$ & $15.62$ & $64.01$ \\
percentage in nocaps  &$0.24$ & $5.05$ & $15.98$ & $35.71$ & $69.91$ \\
\bottomrule
\end{tabular}
\vspace{-2mm}
\caption{Distribution of $568$ object categories on COCO training images and nocaps validation images. Each column is a subset of object categories whose number of occurrences are below the threshold. The percentage is calculated by dividing the counts of those objects by the total counts of all objects in the dataset.}
\label{tab:nocaps_dist}
\end{table}

\begin{figure}[t]
\centering
\includegraphics[width=0.45\textwidth]{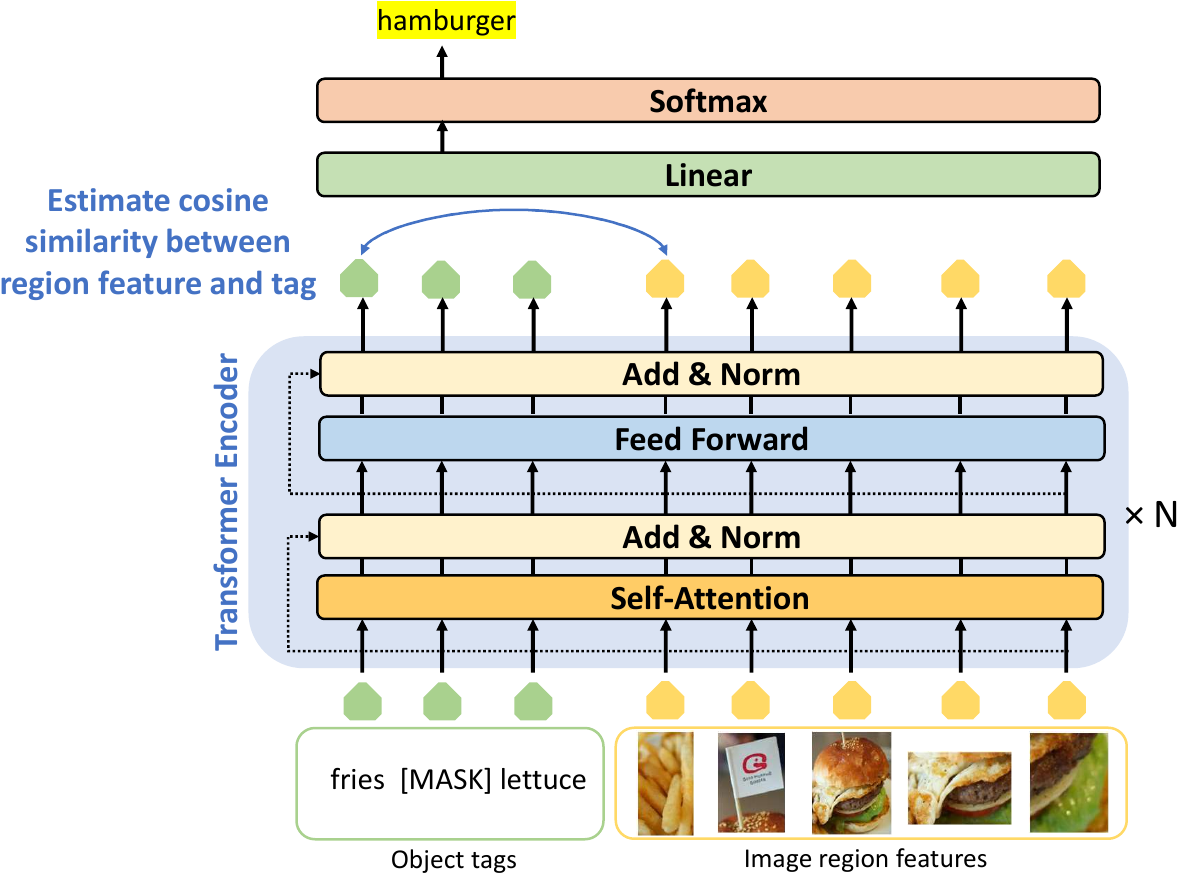}
\vspace{-2mm}
\figcaption{Overview of our VIVO pre-trained Transformer model. Our model consists of multiple Transformer encoder layers followed by a linear layer and a softmax layer. We use masked tag prediction to conduct pre-training. To analyze the visual-text alignment, we use the outputs of the last layer of the encoder layers to estimate the cosine similarity between the image region and tag.}
\label{fig:model_arch}
\figvspace
\end{figure}

\begin{figure*}[t]
\centering
\includegraphics[width=0.9\textwidth]{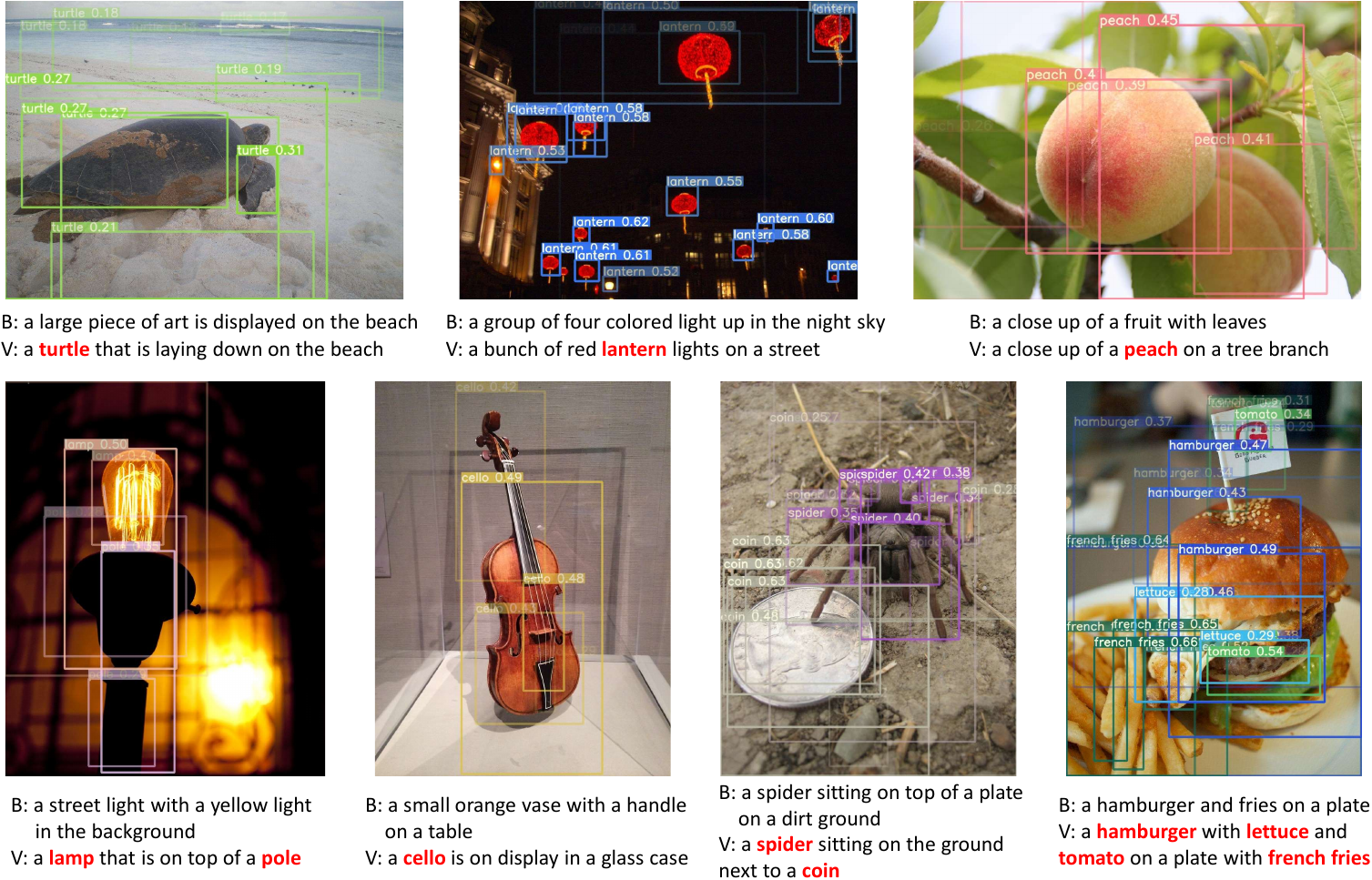}
\vspace{-2mm}
\figcaption{Image captioning results on nocaps. B: our baseline without adding VIVO pre-training. V: our approach with VIVO pre-training. \textcolor{red}{Red} text represents novel objects. % that are presented in the ground-truth object tags.
%and \textcolor{green}{green} text represents novel objects that are not labeled as ground-truth. 
For each image, we show the similarity scores of each image region to the novel objects appear in the captions. The bounding box color is brighter when the similarity is higher.
%We also show the results after adding Constrained Beam Search to B and C in the first 3 images as B+C, V+C, where C is the constraints used for CBS.
} 
\label{fig:nocaps}
\figvspace
\end{figure*}

\subsection{Visual-Text Alignment}
\kevin{To further understand the effects} of VIVO pre-training in learning visual vocabulary, 
%, \ie, aligning
which aligns image regions with object tags, 
we show how the novel object tags can be grounded in image regions in Figure~\ref{fig:nocaps}. 
Given the images from the Open Images validation set, we extract image region features using the same object detector from UpDown and generate captions from the captioning model with VIVO pre-training. 
After identifying the novel objects in the generated captions, \kevin{as shown in Figure~\ref{fig:model_arch}}, we feed the novel object tags, together with the extracted image region features, to the VIVO pre-trained Transformer model. The output of the last encoder layer is used as the contextualized representation of the corresponding input. We then calculate the cosine similarity between representations of each pair of image region and object tag. We highlight the pairs with high scores in Figure~\ref{fig:nocaps}. The result shows that our model can precisely align the mentions of these novel objects in captions with the corresponding image regions. % among many noisy regions.

\begin{table}[t]
\centering
\small
\begin{tabular}{l@{\hspace{2mm}}|c@{\hspace{2mm}}c@{\hspace{2mm}}c@{\hspace{2mm}}c}
\toprule
pre-training  & \kevin{BLEU4} & \kevin{Meteor} & \kevin{CIDEr} & \kevin{SPICE}  \\ \midrule
NO & $33.7$ & $27.9$ & $114.7$ & $21.2$ \\
CC (OSCAR) & $34.8$ & $\bf 28.4$ & $118.2$ & $21.6$\\
CC (OSCAR) + OI (VIVO) & $\bf 34.9$ & $\bf 28.4$ & $\bf 119.8$ & $\bf 21.7$\\
%CC (OSCAR) & $34.7$ & $28.4$ & $117.9$ & $21.6$ \\ 
%CC+OID (OSCAR+VIVO) & $34.7$ & $28.4$ & $118.6$ & $21.8$ \\ 
\bottomrule
\end{tabular}
\vspace{-2mm}
\caption{Evaluation on COCO test set of Karpathy split~\cite{karpathy2015deep}. All results are based on single model with cross-entropy optimization.}
\label{tab:cc_oid}
\end{table}

\subsection{General Image Captioning}
VIVO pre-training does not require 
%removes the requirement of 
the paired image-caption data for model training as in conventional VLP methods. 
It opens up an opportunity to leverage additional data sources to improve image captioning models. 
To demonstrate the effectiveness of VIVO pre-training on general image captioning tasks, 
we trained two versions of OSCAR, following the setting in~\citet{li2020oscar}.
The first OSCAR model is trained solely on
Conceptual Captions (CC)~\cite{sharma2018conceptual}, as described in \citet{li2020oscar}.
The second OSCAR model is pre-trained using VIVO on Open Images (OI), and then fine-tuned on CC. 
%we compare the performance of adding Open Images (OI) to Conceptual Captions (CC)~\cite{sharma2018conceptual} for image captioning. 
%In this joint training, we follow the settings in~\citet{li2020oscar} for the CC dataset and our VIVO pre-training for the Open Images dataset while sharing the same Transformer model. 
As shown in Table~\ref{tab:cc_oid}, VIVO pre-training improves the model performance across all metrics evaluated on the COCO test set, especially in CIDEr score. 
\xiyin{We do observe, however, that the gain on the COCO benchmark is not as substantial as that on the nocaps benchmark. 
We conjecture that this is due to the COCO dataset containing only a small number of visual concepts and thus diminishing the benefit of learning a large visual vocabulary. % cannot be reflected on this evaluation set.
\leizhang{It is also worth noting that using machine-generated image tags rather than human-written captions makes it possible to utilize potentially unlimited amounts of images, which we will pursue in our future work.}}

\begin{table}[ht]
\centering
\small
\begin{tabular}{l@{\hspace{2mm}}|c@{\hspace{2mm}}c@{\hspace{2mm}}c@{\hspace{2mm}}c}
\toprule
\kevin{Tag size} & \kevin{BLEU4} & \kevin{Meteor} & \kevin{CIDEr} & \kevin{SPICE}  \\ \midrule
$0$ (w/o VIVO) & $18.3$ & $24.2$ & $69.6$ & $11.3$ \\
% ${\bf{T}}_b$ & $20.6$ & $25.4$ & $76.5$ & $11.9$  \\ 
% ${\bf{T}}_b$ + ${\bf{T}}_i$ & $\bf 21.2$ & $\bf 25.4$ & $\bf 77.8$ & $\bf 12.0$ \\
$500$ classes & $20.6$ & $25.4$ & $76.5$ & $11.9$  \\ 
$6.4K$ classes & $\bf 21.2$ & $\bf 25.4$ & $\bf 77.8$ & $\bf 12.0$ \\
\bottomrule
\end{tabular}
\vspace{-2mm}
% \tabcaption{Ablation study of VIVO pre-training using different tag sizes. Results are evaluated on the entire validation set of nocaps.}
\tabcaption{Adding VIVO pre-training makes substantial improvement on NOC. Using more labels in pre-training also gives better results. All the models are fine-tuned on COCO and evaluated on the validation set of nocaps.}
\vspace{-2mm}
\label{tab:tag}
\end{table}

\begin{table}[h]
\centering
\small
\begin{tabular}{l@{\hspace{2mm}}|c@{\hspace{2mm}}c@{\hspace{2mm}}c@{\hspace{2mm}}c}
\toprule
Loss & \kevin{BLEU4} & \kevin{Meteor} & \kevin{CIDEr} & \kevin{SPICE}  \\ \midrule
Mask only one token & $20.6$ & $25.2$ & $74.9$ & $11.8$  \\ 
w/o Hungarian matching & $21.0$ & $25.4$ & $75.8$ & $11.8$ \\ 
\kevin{w/} Hungarian matching & $\bf 21.2$ & $\bf 25.4$ & $\bf 77.8$ & $\bf 12.0$ \\
\bottomrule
\end{tabular}
\vspace{-2mm}
\tabcaption{Ablation study of the proposed Hungarian matching loss.
Results are evaluated on the entire validation set of nocaps.
}
\vspace{-4mm}
\label{tab:loss}
\end{table}

\subsection{Ablation Study}
We select a subset of $10\%$ images from the Open Images training set to conduct an ablation study. We fine-tune with cross-entropy loss on the COCO dataset and report the performance on the nocaps validation set.

\Paragraph{Using a Larger Set of Tags} 
We investigate whether % scaling the lexicon of tags 
using a larger set of tags
in pre-training improves performance of the downstream image captioning task.
We select $500$ classes of objects, which are %the classes
used to train the object detector, from the overall $6.4K$ classes of tags to conduct VIVO pre-training.
%We prepare two splits of labels from Open Images: object tags (${\bf{T}}_b$) and image-level tags (${\bf{T}}_i$). There are $500$ classes in ${\bf{T}}_b$ and $6.4K$ classes in ${\bf{T}}_i$. ${\bf{T}}_i$ does not include additional annotations belonging to the $500$ classes in ${\bf{T}}_b$.
As shown in Table~\ref{tab:tag}, VIVO pre-training with $500$ classes significantly improves the performance on nocaps by $6.9\%$ compared to no pre-training. Expanding the labels to $6.4K$ classes can further improve the performance, although the gain is limited due to the increased diversity of objects presented in test images. 

% \Paragraph{Criterion for Set Prediction}
\Paragraph{Using Hungarian Matching Loss}
We evaluate the effectiveness of the proposed Hungarian matching in VIVO pre-training to predict a set of tags. Training without Hungarian matching reduces the tag prediction to 
%means the default 
the standard masked language modeling task, which predicts the masked tokens in the same order as that in the input sequence. 
In addition, we also perform VIVO pre-training by masking only one token in input, which makes word order information not useful.
%but less efficient as to empirical evidence. 
The evaluation results on the nocaps validation set are in Table~\ref{tab:loss}. We can see that masking only one token is not effective, and using Hungarian matching leads to the best model performance. 
%is effective to handle the unordered tags in VIVO pre-training. 

\section{Conclusions}

We have presented a weakly supervised learning approach to training image captioning models in two steps. First, a Transformer-based model is pre-trained on large amounts of image-tag pairs to learn a visual vocabulary without the need of using image-caption pairs which are harder to obtain.
Then, the model is fine-tuned on image-caption pairs to learn to incorporate information from the pre-trained visual vocabulary and compose
%, for a given image, 
%a caption template, which can be filled with object tags to form 
image captions that can describe novel visual objects unseen in the training data of image-caption pairs.
% which are used for model fine-tuning. 

Our experiments on the nocaps benchmark dataset demonstrate that our model achieves compositional generalization, allowing for zero-shot generalization to novel objects for image captioning.  
As a result, our best single model creates new state-of-the-art that surpasses the human CIDEr score on nocaps.
A detailed analysis reveals that the generalization is attributed to a large degree to the visual vocabulary learned in model pre-training, which maps visual objects or regions with similar semantic meanings to feature vectors that are close to each other in a discrete semantic space.

Since our pre-training does not need paired image-caption data, one of our future works is to leverage large amounts of vision data, beyond image-tag pairs used in this paper, to significantly improve the quality of the visual vocabulary.  

%In this paper, we have presented VIVO pre-training for novel object captioning.
%As the first VLP method that does not rely on paired image-sentence data, \kevin{VIVO leverages large amounts of 
% -scale vision dataset with image-tag pairs in pre-training to learn cross-modality alignment, and thus builds a rich visual vocabulary for novel object captioning.} We show that the tags of novel objects can be aligned with the corresponding image regions using visual vocabulary learned in VIVO pre-training.

% grounded in image regions with the alignment learned from VIVO pre-training.
%As the first vision-language pre-training method that does not rely on paired image-sentence data, VIVO can leverage data source with image-tag pairs to conduct pre-training at the sub-sentence level.
% \kevin{We have studied} various aspects in VIVO pre-training including tag sizes and loss functions to tackle the unordered set of tags. 
% The visual vocabulary, or the visual-text alignment, can be quantitatively evaluated via our proposed visual-text alignment analysis. Our method has surpassed human performance on the nocaps benchmark. 
% \kevin{The proposed method achieves new state-of-the-art results on the nocaps benchmark, and surpasses the human CIDEr score.}

\section{Acknowledgements}
We thank Jianfeng Wang, Ehsan Azarnasab, Lin Liang, Pengchuan Zhang, Xiujun Li, Chunyuan Li, Jianwei Yang,  Yu Wang, Houdong Hu, Furu Wei, Dong Li for valuable discussions and comments.

% \begin{small}
\bibliography{egbib}
% \end{small}
\appendix
% \begin{center}\large\bf
% SUPPLEMENTARY MATERIAL
% \end{center}
\vspace{\baselineskip}
\input{nocaps_06_supp}

\end{document}

%% file: nocaps_06_supp.tex
\twocolumn[{%
\renewcommand\twocolumn[1][]{#1}%
% \maketitle
\begin{center}
\textbf{\large APPENDICES}
\end{center}
\vspace{\baselineskip}
% \begin{figure*}[H]
% \vspace{-50}
\begin{center}
	\setlength{\tabcolsep}{0.0pt}
	\renewcommand{\arraystretch}{2}
	\begin{tabular}{rccc}
	& Baseline (Random initialization) & Baseline (BERT initialization) & VIVO \vspace{-5pt}\\
	\rotatebox[origin=c]{90}{\small{Pre-training}} &
	\raisebox{-.5\height}{\includegraphics[trim=80 80 80 80,clip,height=.33\textwidth]{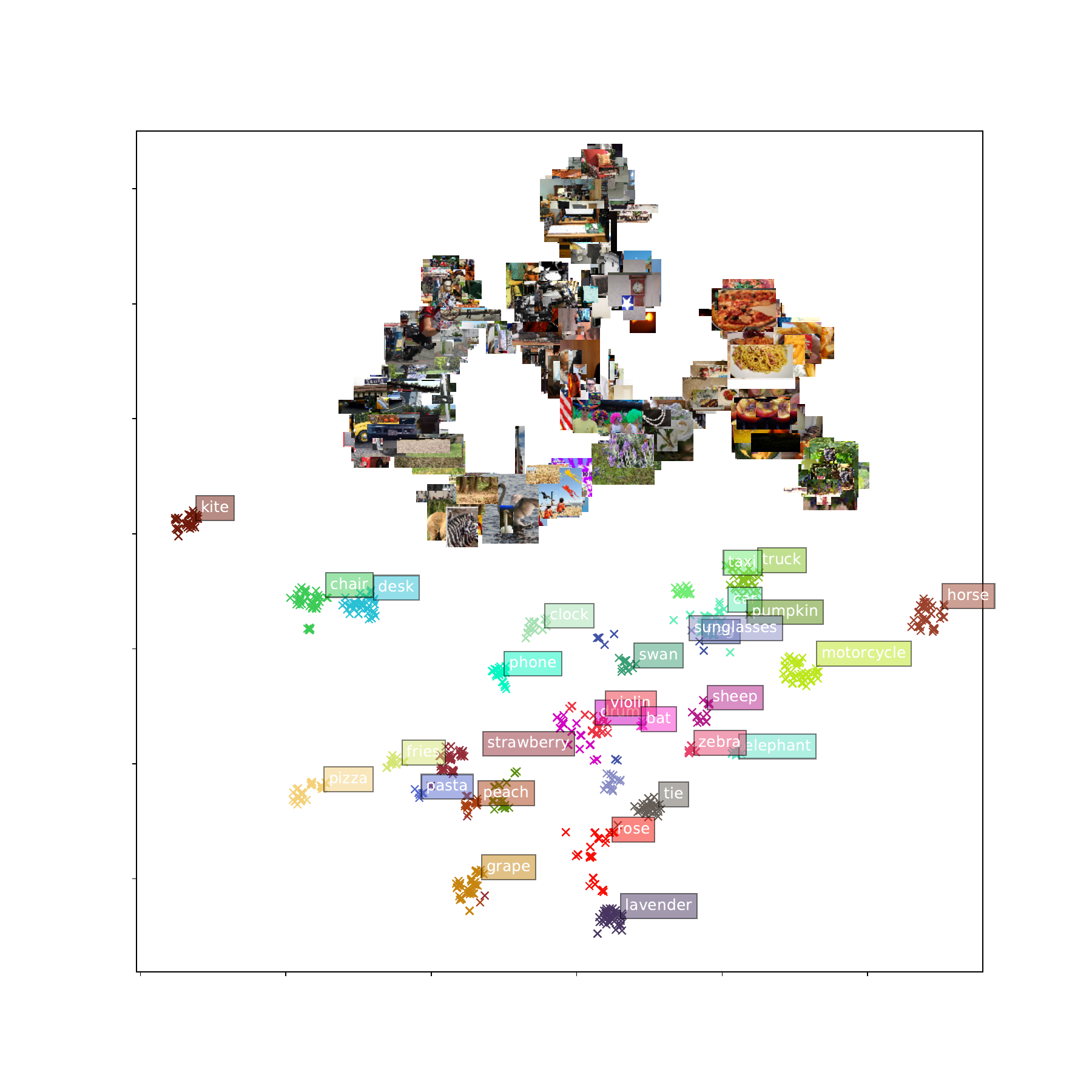}} &
	\raisebox{-.5\height}{\includegraphics[trim=80 80 80 80,clip,height=.33\textwidth]{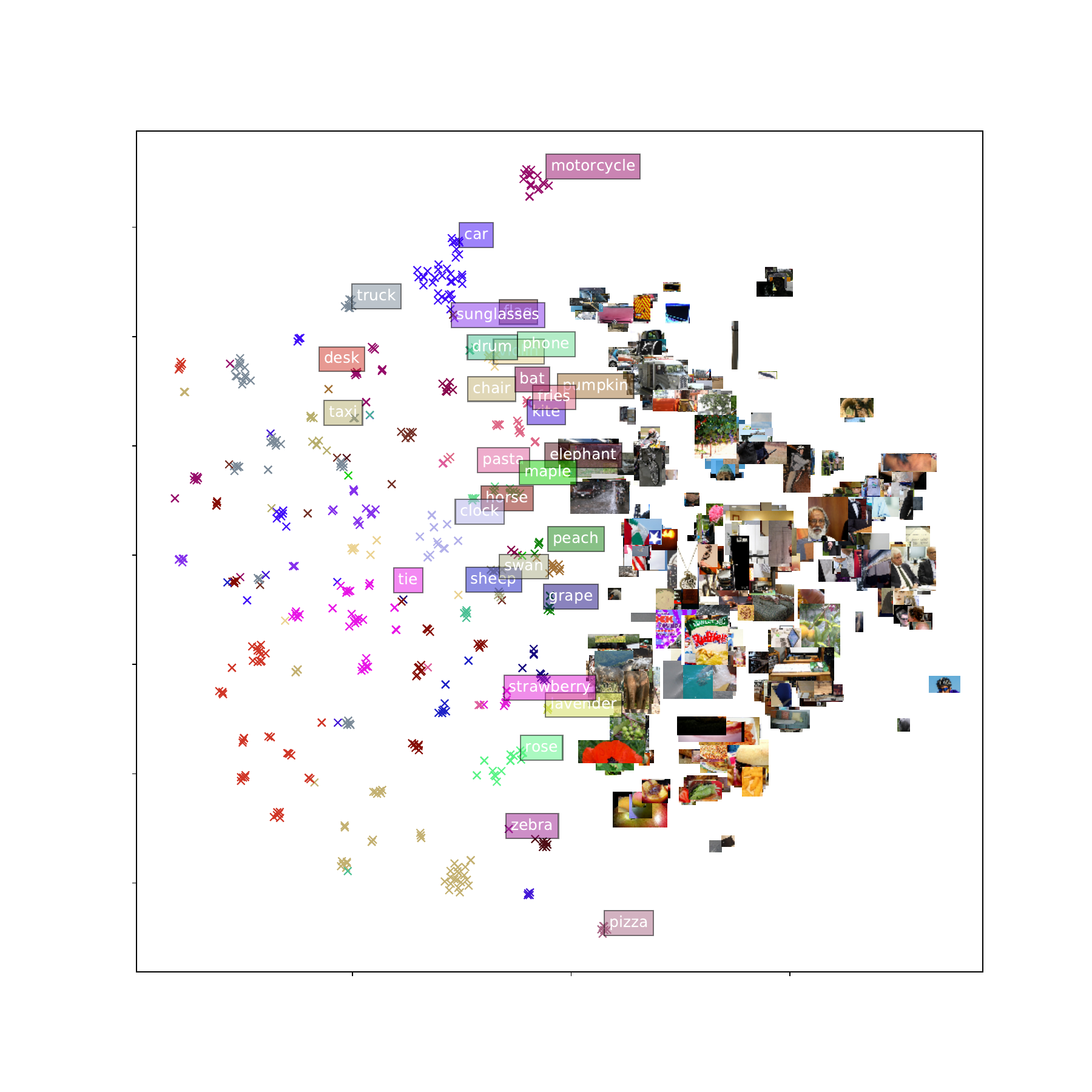}} &
	\raisebox{-.5\height}{\includegraphics[trim=80 80 80 80,clip,height=.33\textwidth]{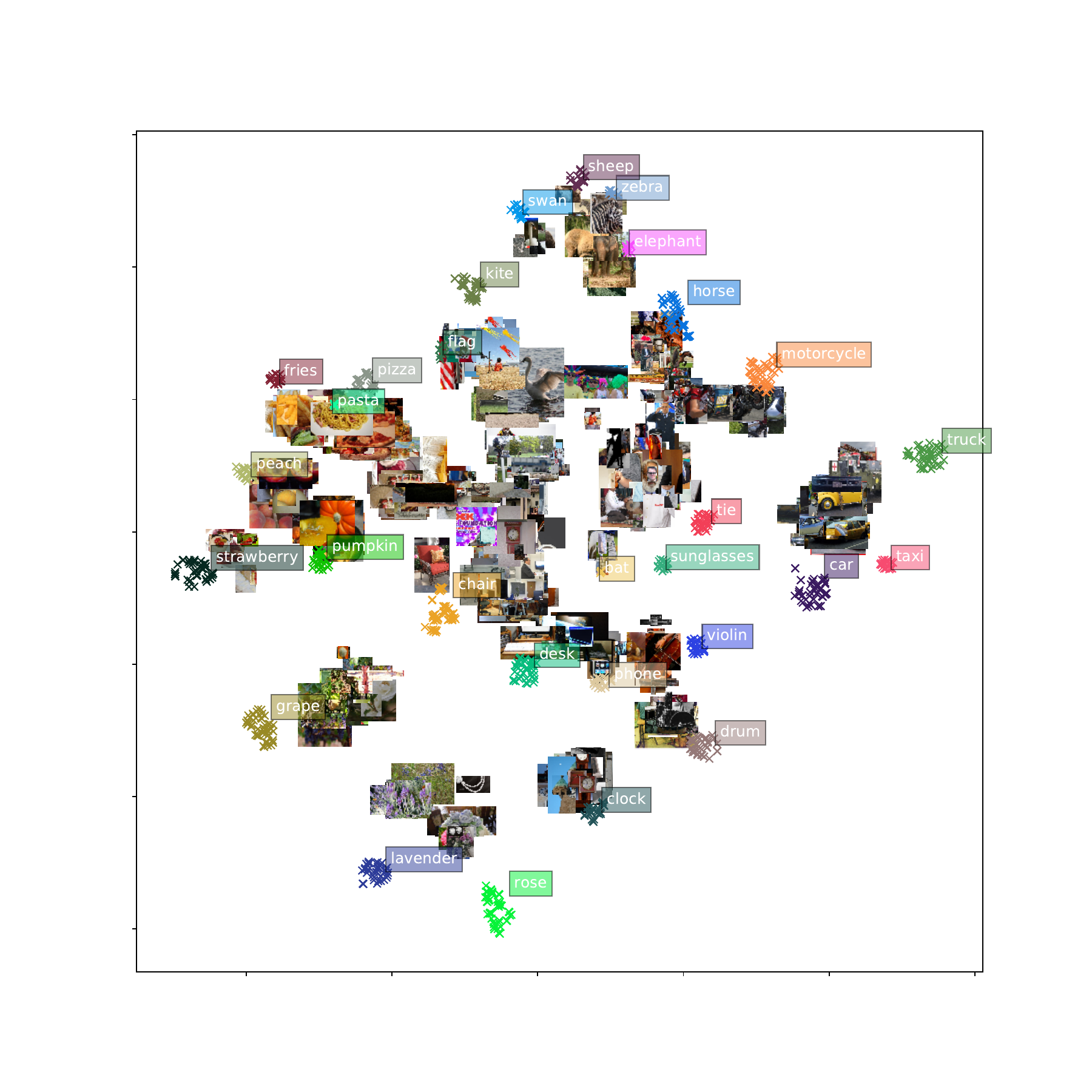}}\vspace{-5pt}\\
	& (a) & (b) & (c)\vspace{-5pt}\\
	\rotatebox[origin=c]{90}{\small{Fine-tuning}} &
	\raisebox{-.5\height}{\includegraphics[trim=80 80 80 80,clip,height=.33\textwidth]{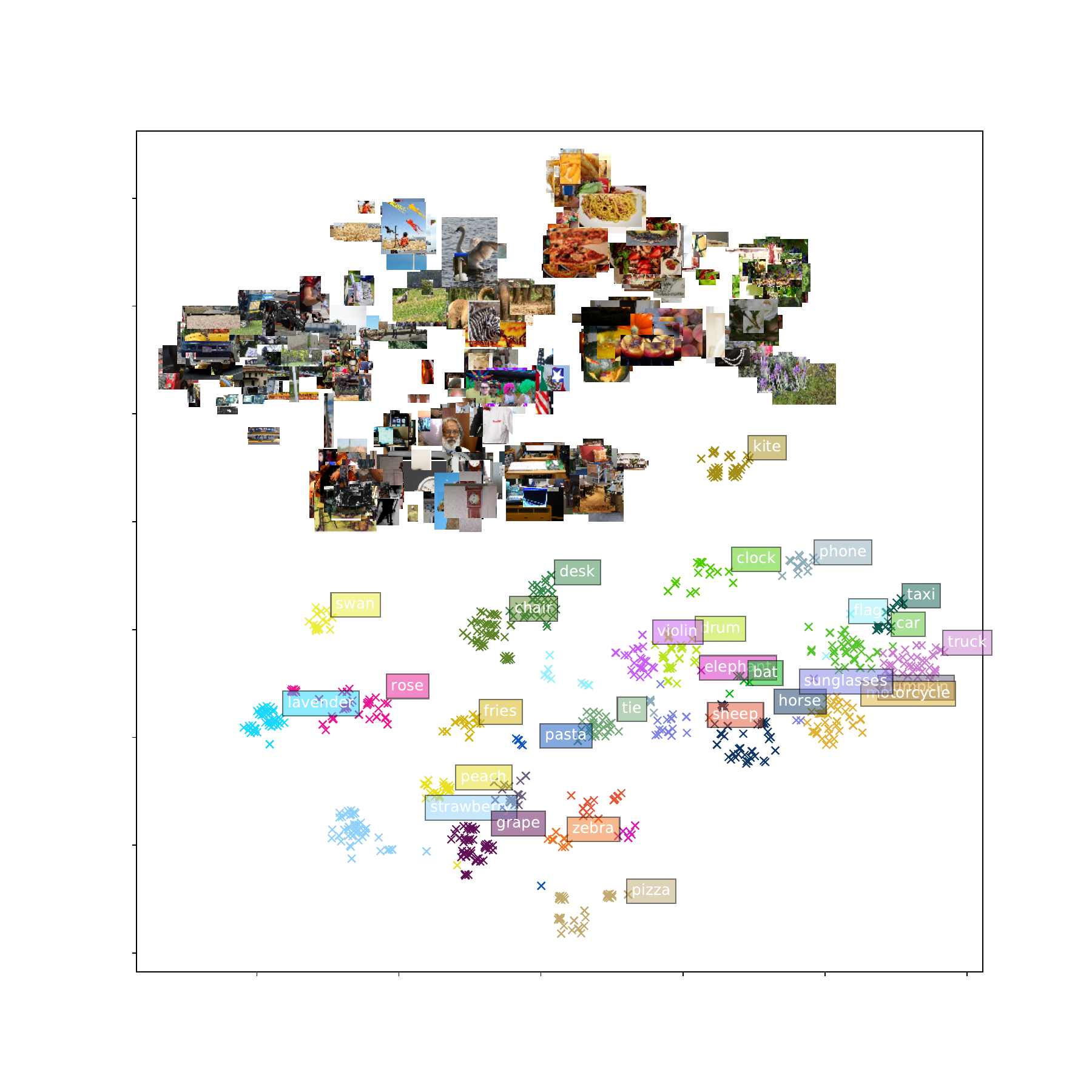}} &
	\raisebox{-.5\height}{\includegraphics[trim=80 80 80 80,clip,height=.33\textwidth]{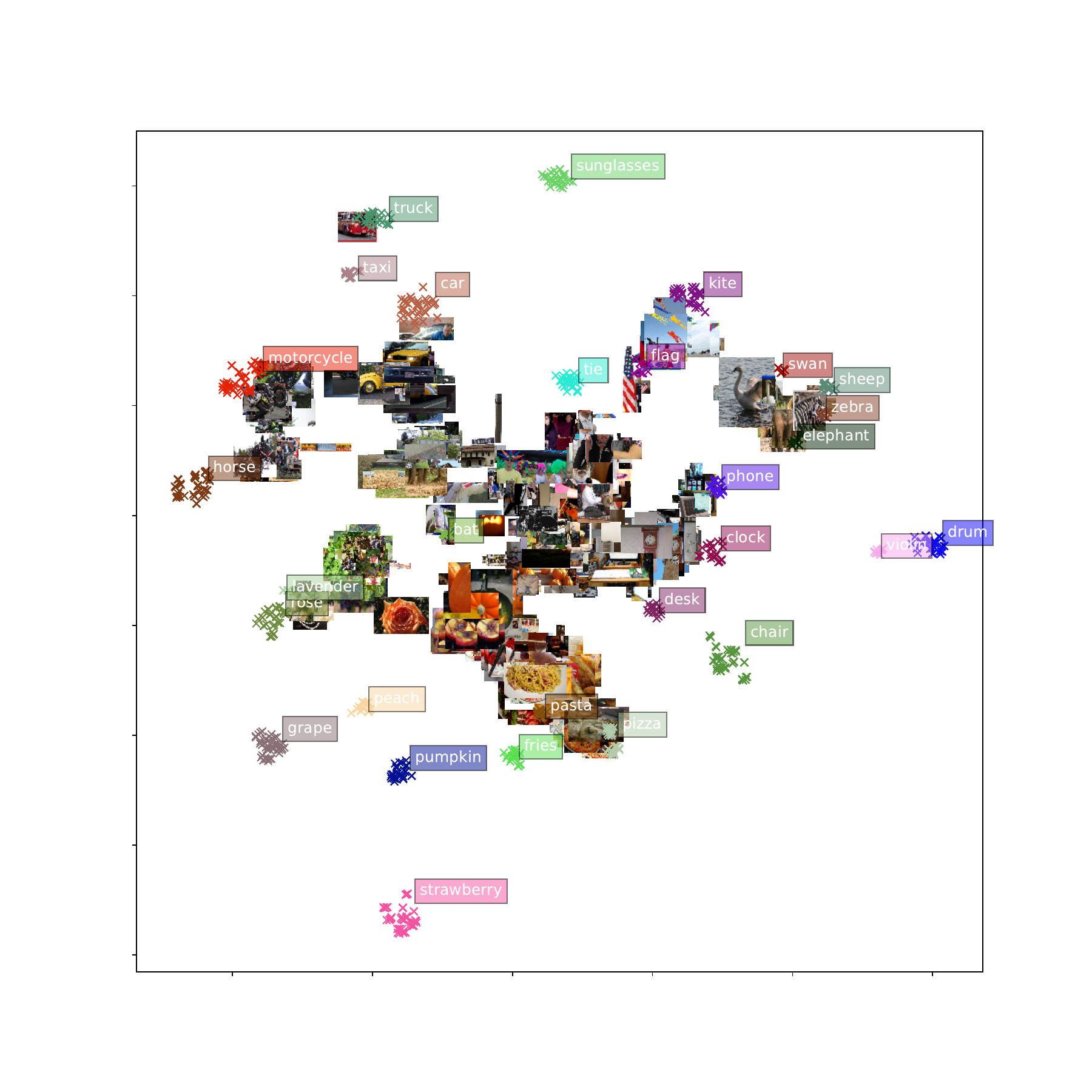}} &
	\raisebox{-.5\height}{\includegraphics[trim=80 80 80 80,clip,height=.33\textwidth]{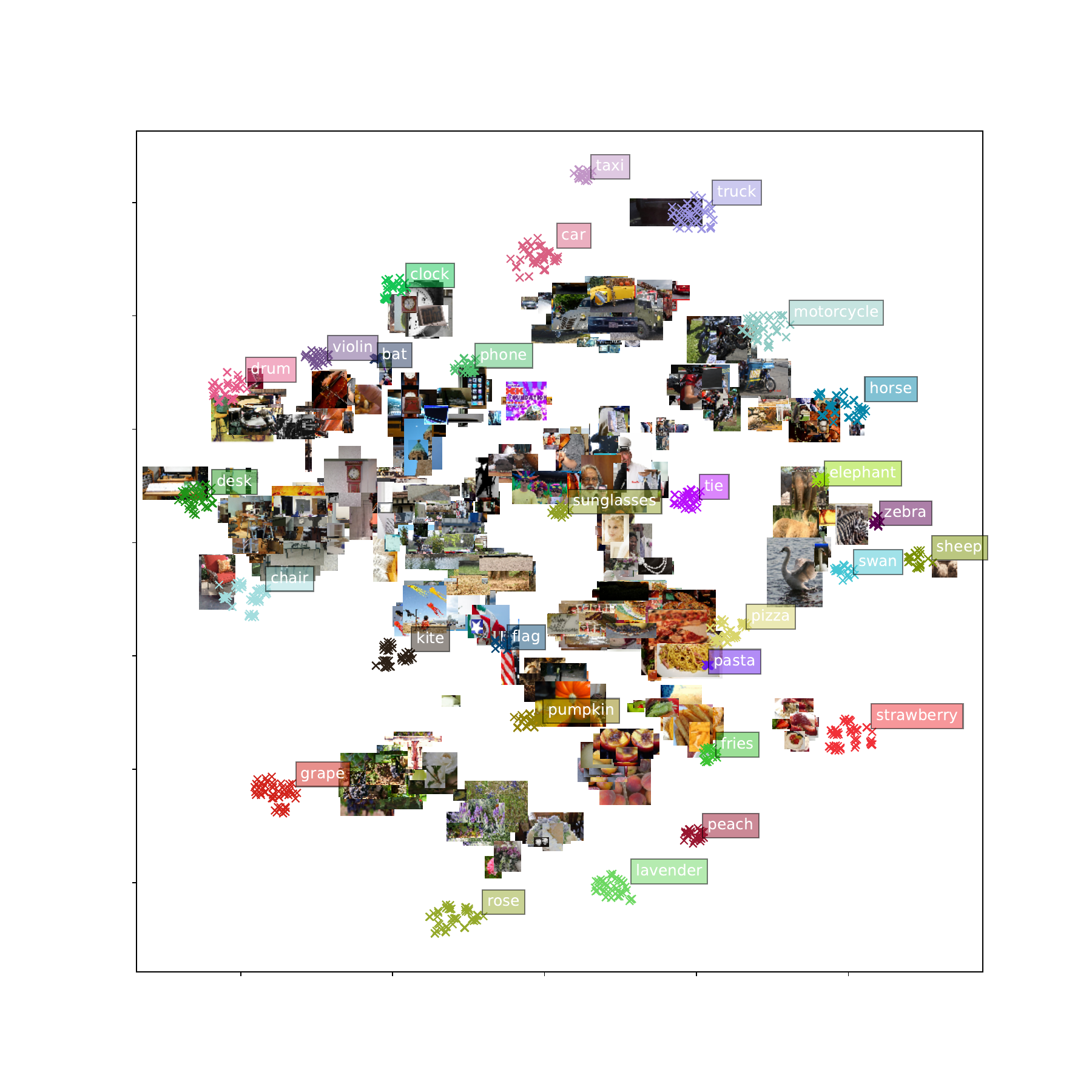}}\vspace{-5pt}\\
	& (d) & (e) & (f)
	\end{tabular}
	\vspace{-15pt}
\end{center}
\figcaption{Feature space visualization results of the baselines and VIVO using $t$-SNE. In each figure, we use an image patch to represent its image feature, and use a marker ``$\times$'' with the same color to indicate the same object tag. As shown in (a) and (d), the baseline with random initialization does not work well for visual-text alignment. In (b), the baseline with BERT initialization does not align the two modalities at first, but the alignment is improved after fine-tuning, as shown in (e). In contrast, our approach improves the visual-text alignment in both pre-training and fine-tuning, as shown in (c) and (f).}
\vspace{10pt}
\label{fig:tsne_comparison}
% \end{figure*}
}]

\section{Visual Vocabulary Visualization}

To further understand the effects of VIVO, we conduct a qualitative comparison between the feature spaces learnt from the baselines and VIVO using $t$-SNE~\cite{maaten2008visualizing}. We randomly sample $30$ object categories from the nocaps validation set, and visualize the representations of the image regions and object tags. % We feed each image region and object tag to the Transformer model, and extract the representations from the last layer of Transformer encoder for visualization. 

Figure~\ref{fig:tsne_comparison} shows the comparison of two baselines and VIVO. %\hl{In each figure, we use an image patch to represent its image feature, and use a marker ``$\times$'' with the same color to indicate the same object tag.} \leizhang{I suggest moving this sentence to the caption of Figure 1 to make it easier to understand the figure.} 
The results show that VIVO compares favorably with the baselines in visual-text alignment. 

\begin{figure*}[t]
\begin{center}
\includegraphics[trim=100 80 90 100, width=1.\textwidth]{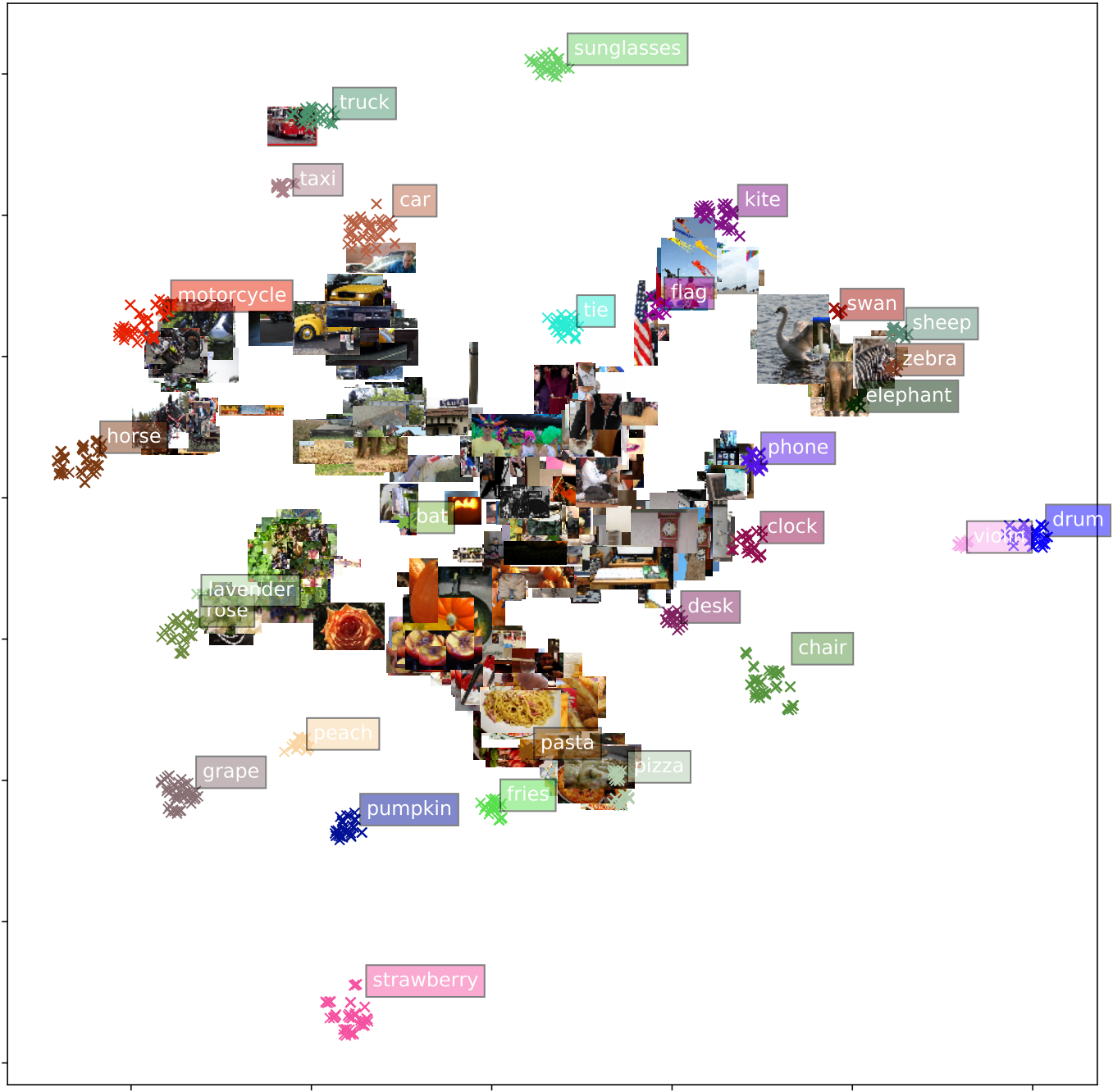}
\end{center}
\figcaption{$t$-SNE visualization of the baseline with BERT initialization and fine-tuning, as shown in Figure~\ref{fig:tsne_comparison}(e). The marker ``$\times$'' with the same color indicates the same object class. We observe that the alignment is better for the objects commonly presenting in the caption corpora, \textit{e.g.,} pizza, motorcycle, but worse for novel objects, \textit{e.g.,} grape, violin, drum, strawberry.}
\label{fig:tsne_alignment_baseline_ft}
\end{figure*}

\begin{figure*}[t]
\begin{center}
\includegraphics[trim=100 80 90 100, width=1.\textwidth]{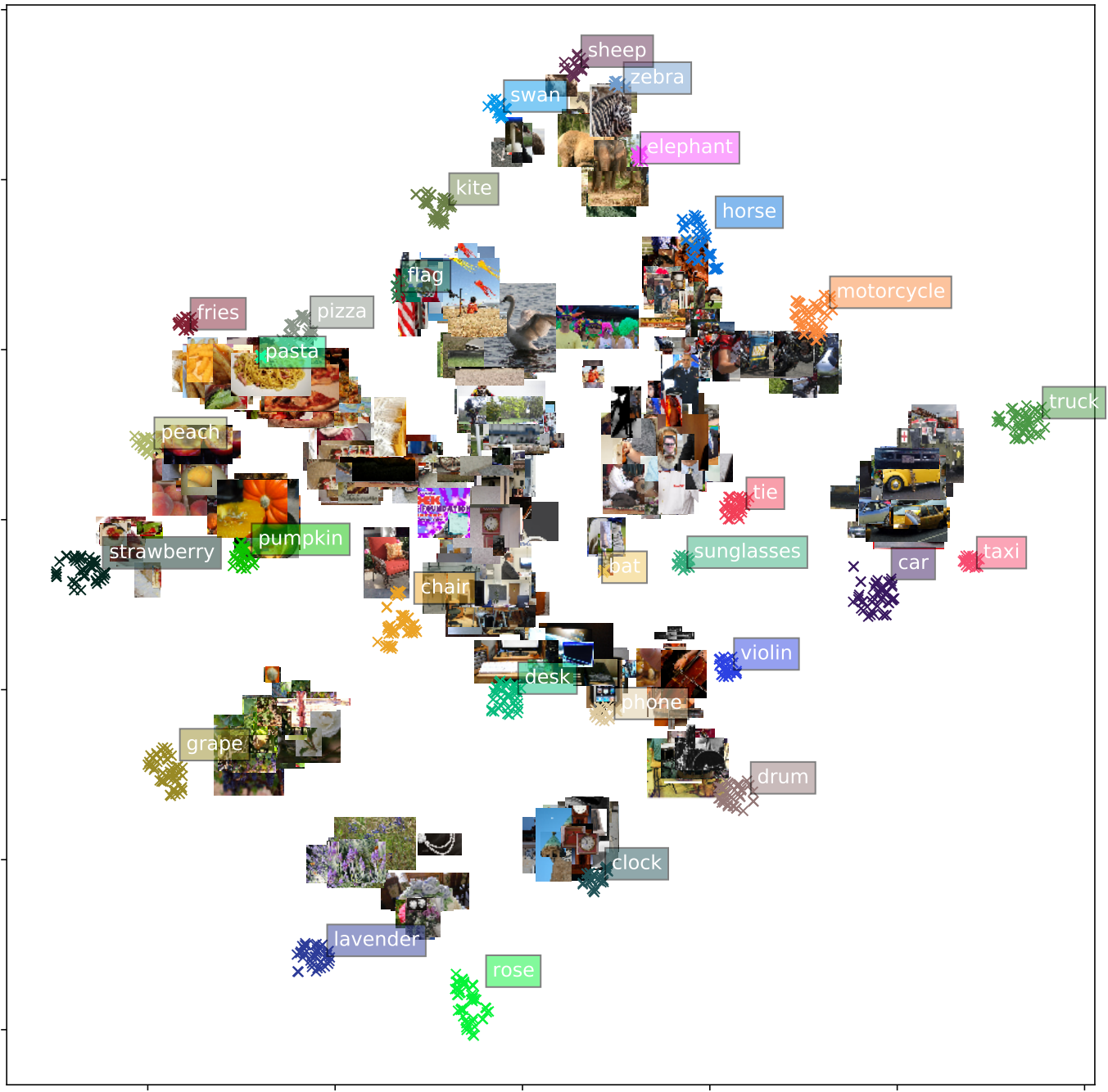}
\end{center}
\figcaption{$t$-SNE visualization of the VIVO pre-trained model, as shown in Figure~\ref{fig:tsne_comparison}(c). The marker ``$\times$'' with the same color indicates the same object class. With the help of VIVO pre-training, we see that the image region features and object tags are better aligned compared to the baselines. }
\label{fig:tsne_alignment_VIVO}
\end{figure*}

\begin{figure*}[t]
\begin{center}
\includegraphics[trim=100 80 90 100, width=1.\textwidth]{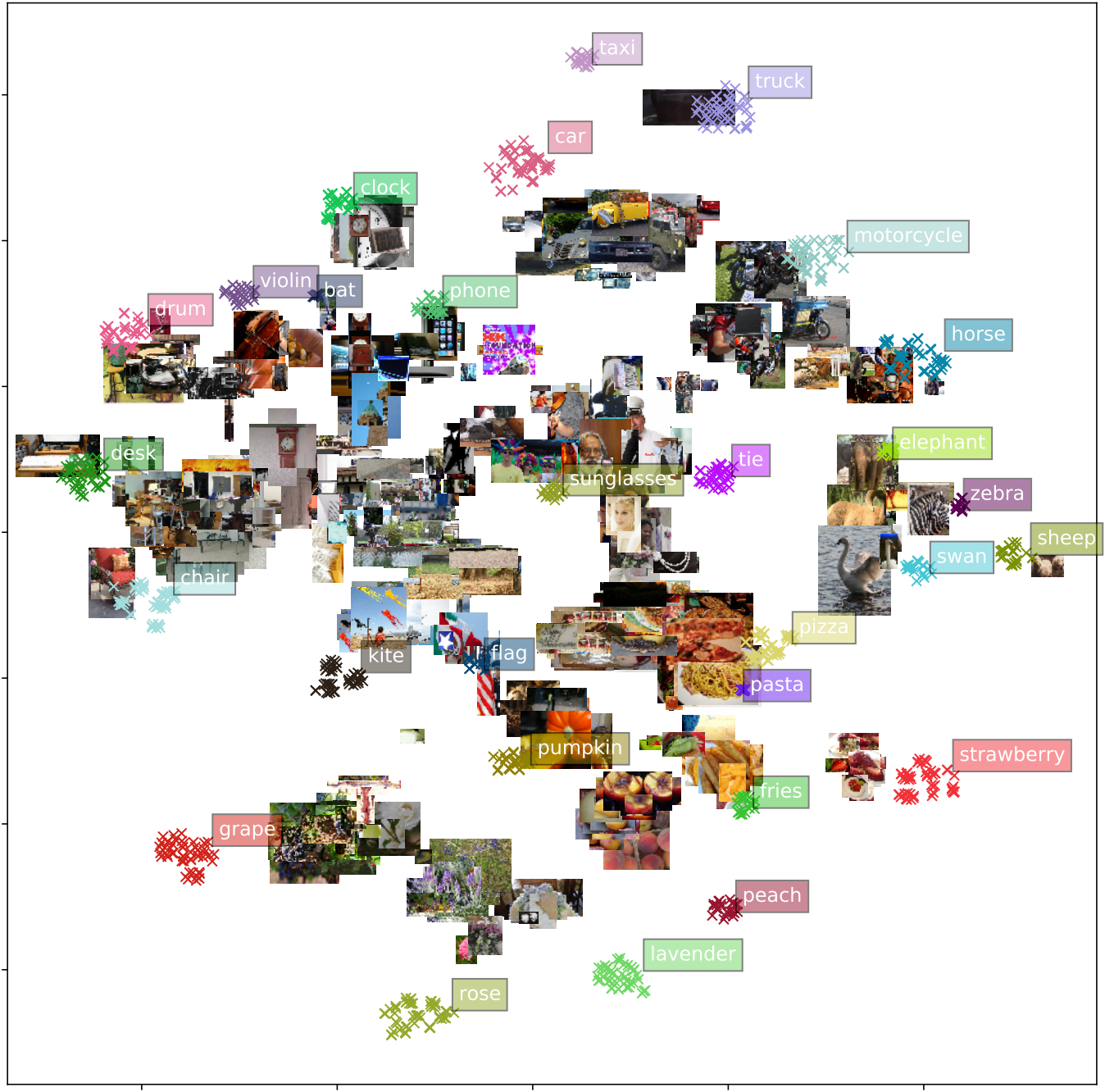}
\end{center}
\figcaption{$t$-SNE visualization of VIVO fine-tuned model, as shown in Figure~\ref{fig:tsne_comparison}(f). The marker ``$\times$'' with the same color indicates the same object class. Our model improves the visual-text alignment overall, especially for novel objects.}
\label{fig:tsne_alignment_VIVO_ft}
\end{figure*}

We enlarge the $t$-SNE visualization results of Figure~\ref{fig:tsne_comparison}(e), Figure~\ref{fig:tsne_comparison}(c), and Figure~\ref{fig:tsne_comparison}(f) in Figure~\ref{fig:tsne_alignment_baseline_ft}, Figure~\ref{fig:tsne_alignment_VIVO}, and Figure~\ref{fig:tsne_alignment_VIVO_ft}, respectively. The results reveal some interesting findings: (\textit{i}) We observe that VIVO pre-training is helpful in learning a better cross-modality alignment compared to the baselines. (\textit{ii}) Fine-tuning with paired image-caption training data can further improve the alignment between two modalities. (\textit{iii}) In Figure~\ref{fig:tsne_comparison}(e), the alignment of the baseline is better for that objects that frequently occur in the caption corpora, \textit{e.g.}, motorcycle, pizza, but worse for novel objects, \textit{e.g.}, violin, drum, grape. (\textit{iv}) VIVO improves the alignment overall, especially for novel objects.

%Figure~\ref{fig:tsne_alignment_baseline} and Figure~\ref{fig:tsne_alignment_VIVO} show the results of the baseline and our pre-trained model, respectively. 
%For each figure, we use the image patch to represent its image feature. We use marker ``$\times$'' to indicate object tag. 
%The results show that VIVO pre-training learns to align the image region features and the object tags correctly. 

%Since we fine-tune the pre-trained model on image captioning task, one may wonder what is the effect on the feature space after fine-tuning. Figure~\ref{fig:tsne_alignment_baseline_ft} and Figure~\ref{fig:tsne_alignment_VIVO_ft} show the results of the fine-tuned baseline and our fine-tuned model, respectively. The results reveal some interesting findings: (\textit{i}) We observe that both models are learning the visual-text alignment during fine-tuning. (\textit{ii}) For the baseline model, the alignment is better for the objects frequently present in the caption corpora, \textit{e.g.}, motorcycle, pizza, but worse for novel objects, \textit{e.g.}, violin, drum, grape. (\textit{iii}) Our proposed model improves the alignment overall, especially on novel objects.

% With the help of VIVO pre-training, the distance of the same object between two modalities is substantially reduced compared to the baseline. This validates the importance of VIVO pre-training for image captioning. 

\section{Implementation Details}
Our transformer-based image captioning model consists of $12$ transformer layers for encoding and a linear layer for prediction. Note that the model does not have any decoder layer. We use WordPiece embedding~\cite{wu2016google} with a $30,000$ token vocabulary to represent input words, including both object tags and captions. For a given token, its input representation is constructed by summing the corresponding token, segment, and position embeddings.
In addition, we use the Faster R-CNN model from UpDown~\cite{anderson2018bottom} to extract image region features and a tagging model trained on the Open Images dataset to predict tags.
The transformer model is first pre-trained then fine-tuned, and applied iteratively at inference time to generate the output. % a sequential output.

\Paragraph{Pre-training} As described in the main text of the paper, our model consumes a set of tags as textual inputs during pre-training. In addition to the ground truth labels from the Open Images training set, we also use the predictions from our tagging model to enhance the label quality to mitigate that the labels in the Open Images dataset are not complete. We tokenize the tags, and concatenate the tokens into a sequence. We also add the special token [SEP] at the end of the sequence. Following masked language modeling of BERT, we randomly choose $15\%$ of tokens for prediction, \ie, replacing the chosen token with (1) the [MASK] token $80\%$ of the time (2) a random token $10\%$ of the time (3) the unchanged token $10\%$ of the time. We concatenate the textual feature sequence and the visual feature sequence to form the input to the model. %, and use the result as input to the model.

\Paragraph{Fine-tuning} The textual input encompasses a caption sentence (i.e., the ground truth of COCO Captions), and a set of tags (i.e., the prediction of the tagging model). 
The sequence for the caption always starts with the [CLS] token and ends with the [SEP] token. The sequence for tags is constructed in the same way as described in pre-training. To differentiate the caption from tags, we add a learned segment embedding to every token indicating whether it belongs to the caption or the tag sequence. In fine-tuning, we only mask tokens from the caption for prediction. The caption feature sequence, tag feature sequence and visual feature sequence are concatenated and fed into the model.

\Paragraph{Inference} At inference time, the model's input contains three parts: a previous prediction for caption, a set of predicted tags, and image region features. At the beginning, the caption part is a [CLS] token followed by a [MASK] token. We feed the input made up of three parts to the model and get the prediction at the position of the [MASK] token. In the next step, we replace the previous [MASK] token with the prediction, and insert another [MASK] token at the end of the caption sequence. This step iterates until the prediction of the end of sentence token, \ie, the [SEP] token, or reaching the maximum length. In this way, the model generates a caption sentence from left to right.